\documentclass[sigconf]{acmart}

\def\BibTeX{{\rm B\kern-.05em{\sc i\kern-.025em b}\kern-.08emT\kern-.1667em\lower.7ex\hbox{E}\kern-.125emX}}

\usepackage{lipsum}
\usepackage[caption=false]{subfig}
\usepackage[ruled]{algorithm2e} 
\SetKwComment{Comment}{$\triangleright$\ }{}
\usepackage{bm}
\usepackage{multirow}

\newcommand{\rulesep}{\unskip\ \vrule\ }

\renewcommand\footnotetextcopyrightpermission[1]{} 
    
\begin{document}

%
\title{Learning Indoor Layouts from Simple Point-Clouds}

%



\author{Md. Tareq Mahmood}
\affiliation{%
 \institution{Bangladesh University of Engineering and Technology}
 \city{Dhaka}
 \country{Bangladesh}}
\email{tareqmahmood@cse.buet.ac.bd}

\author{Mohammed Eunus Ali}
\affiliation{%
 \institution{Bangladesh University of Engineering and Technology}
 \city{Dhaka}
 \country{Bangladesh}}
\email{eunus@cse.buet.ac.bd}
 




%

\renewcommand{\shortauthors}{Mahmood and Ali}

%
\begin{abstract}
Reconstructing a layout of indoor spaces has been a crucial part of growing indoor location based services. One of the key challenges in the proliferation of indoor location based services is the unavailability of indoor spatial maps due to the complex nature of capturing an indoor space model (e.g., floor plan) of an existing building. In this paper, we propose a system to automatically generate floor plans that can recognize rooms from the point-clouds obtained through smartphones like Google's Tango. In particular, we propose two approaches - a Recurrent Neural Network based approach using Pointer Network and a Convolutional Neural Network based approach using Mask-RCNN to identify rooms (and thereby floor plans) from point-clouds. Experimental results on different datasets demonstrate approximately 0.80-0.90 Intersection-over-Union scores, which show that our models can effectively identify the rooms and regenerate the shapes of the rooms in heterogeneous environment.
\end{abstract}

%
%


%

%
\maketitle
\pagestyle{plain}

\section{Introduction}

 With the recent development of smartphone based technologies, indoor location based services \cite{chon2011lifemap, liu2012iparking, ran2004drishti, han2014building} are increasingly attracting attention from both academia and industry in recent years due to their diversified applications. These indoor applications include navigating inside a shopping mall or an airport, assisting a visually challenged person to navigate on his own, enabling first responders to track down a person in emergency inside an unknown building, and helping tourists to navigate inside unfamiliar museums and art galleries. The detailed spatial map (i.e., floor plans) of the indoor space is a prerequisite for enjoying the benefits of these services. Though, indoor maps are now available for some large installations like airports or shopping complexes, the detailed indoor maps are not readily available for most of the small and medium sized indoor spaces such as home and commercial office spaces. One of the main reasons for the unavailability of the detailed maps of these indoor spaces is the  absence of cheap and ubiquitous technologies for re-creating the indoor maps of existing buildings. In this paper, we propose a smartphone assisted solution that exploits deep learning \cite{lecun2015deep} to automatically generate the detailed layout of the floor plan from the point-clouds obtained from a simple scan by a mobile phone like Google's Tango device.

\begin{figure}[t!]
    \centering
        \subfloat[Raw Point-cloud]{\includegraphics[width=0.48\columnwidth]{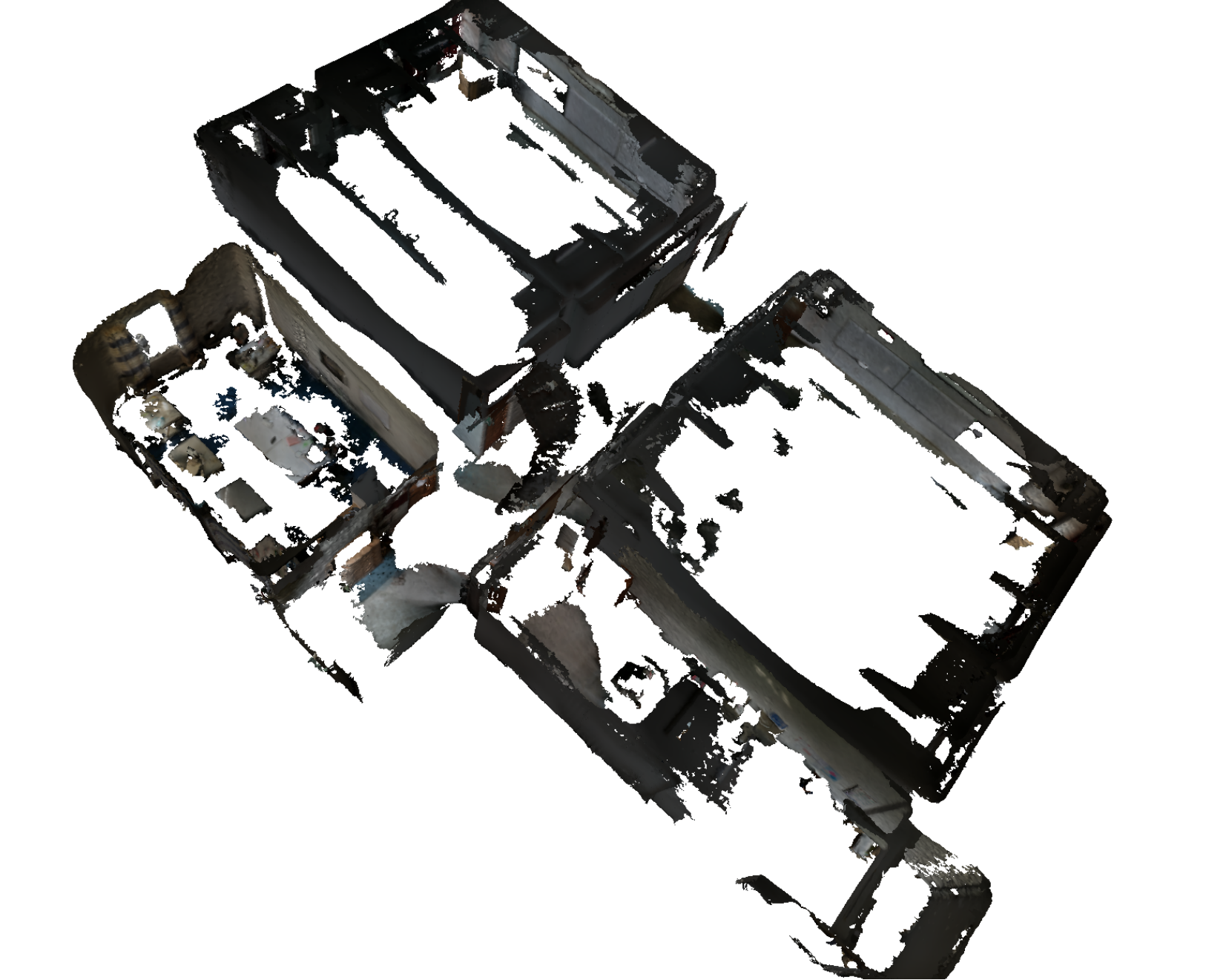}}
        \rulesep
        \subfloat[Indoor Layout]{\includegraphics[width=0.48\columnwidth]{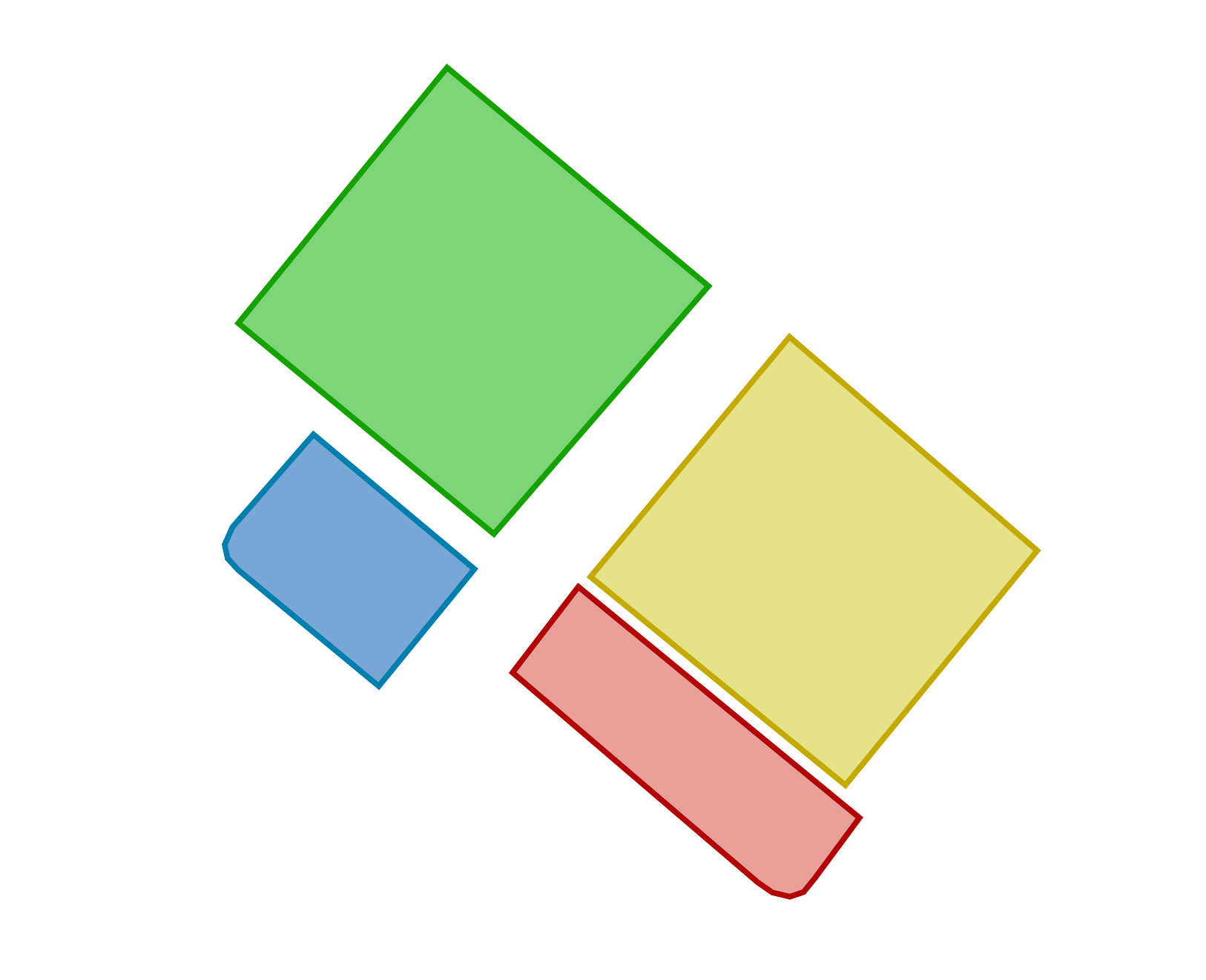}}
    \caption{Extracting indoor layout from point-clouds. Left: the point-cloud collected from an indoor scene. Right: rooms extracted from the scene.}
    \label{fig:intro}
\end{figure}


Traditionally, re-creating floor-plans require the data-acquisition of the indoor spaces through laser sensors \cite{vosselman2014design, jung2017automatic}, sonar sensors \cite{tardos2002robust} or smartphone video \cite{pintore2014interactive,pintore2014effective}, as the first step, and then generation of detailed floor plans of building through human intervention. Recently, some researchers have used expensive heavyweight equipment like LiDAR to capture point-clouds \cite{vosselman2014design, jung2017automatic} of an indoor space, and then develop different algorithms to automatically identify different objects. Jung et al. \cite{jung2017automatic} propose an automatic room detection system from the \emph{complete} point-clouds of the space. The key idea of their approach is to first select floors and ceilings points, then select the points near the ceiling and finally project those points to find a layout. The major limitation of this approach is as follows: they require the complete point-clouds that include floor and ceiling of the space, which is hard to capture. Thus if a portion of ceiling or floor is missing in the point-cloud, this approach fails to remove background clutters and thereby cannot detect rooms in such scenarios. Similarly, Armeni et al. \cite{armeni20163d} propose a system that first detects the vertical planes of the walls, and then divides the entire scene into multiple 3D spaces based on the detected vertical planes, and finally merges adjoining 3D spaces if there is no wall between them. The major limitation of this approach is as follows. They assume that the walls are aligned with X and Y axis (i.e., Manhattan Assumption) and hence fails to detect \emph{rooms with acute or obtuse angles or curved walls}. More recently, a neural network architecture, FloorNet \cite{liu2018floornet} has been proposed that converts an RGBD video stream of a large 3D space into floor-plan. They have used both point clouds and RGB images to generate the floor-plan. The major limitation of their approach is that they fail to generate layout when the point-clouds generated from the scan is incomplete or noisy (e.g., large gaps for windows, missing a portion of a wall, furniture, etc.), which is the case for point-clouds generated by a simple scan using a hand-held device like Google's Tango. 

To alleviate the above limitations, in this paper, we propose a comprehensive deep learning based room detection and indoor layout generation technique. One may assume that simple geometric techniques like \textit{convex hull} or image processing techniques like \textit{contour detection} can be used to detect rooms and generate layout from point-clouds. However, these have following major challenges. (i) Though convex hull algorithm can be used to detect the boundary of a single room, in a multi-room scenario, convex hull ends up in detecting boundary of all the rooms combined. (ii) The contour detection works when each room has closed boundary, whereas often point-cloud of a scene will have gaps, and contour detection technique may merge multiple rooms into one room.

In this paper, we propose two deep learning based techniques: the first  method is based on the concept of geometric shape learning, and the second one is based on image segmentation network, to automatically build indoor maps showing locations of rooms from noisy and incomplete point-clouds obtained from the scanning of the physical space by modern smartphones. The key intuition of our geometric learning approach comes from the proposed Pointer Network \cite{vinyals2015pointer} that was originally proposed to learn triangulation and convex hull from a given set of points. We adopt the Pointer Network architecture to learn rooms in terms of sequence of connected significant border points. On the other hand, the key motivation of exploiting the image segmentation network, namely Mask R-CNN \cite{he2017mask}, to identify rooms and layouts, comes from the recent success in object identification from image using Mask R-CNN \cite{he2017mask}. One of the main challenges in applying these deep learning techniques for our problem is the scarcity of 3D indoor scene data. To tackle this problem, we introduce a novel approach towards generating a synthetic indoor dataset to train deep learning models, which shows satisfactory performance in real-world evaluation.


Our approach works in two steps: In the first step, we extract boundary candidate points from point-clouds, which essentially depicts different walls of the indoor space. In the next step, we build deep learning network, i.e., PtrNet or Mask R-CNN, to learn rooms from these candidate point sets.

To detect walls from from point-clouds, we adapt the histogram projection based technique, which is a variant of the technique proposed by Adan et al. \cite{adan20113d}. We first project horizontal coordinate of each point onto the floor to get a 2D histogram. If there is a wall on a bin of the 2D histogram, it usually has more points than those bins that do not have any wall on them. We introduce a dynamic and robust threshold derived from the data. We keep only the bins containing points greater than the threshold. This gives us the positions of walls in the 2D histogram as an image. 

As the second step, to identify rooms from the candidate point sets, we build two neural network models to learn individual rooms from the 2D histogram of point-clouds. The key intuition of using PtrNet to identify rooms is to learn significant point set that determine the shape of each room from a given sequence of point clouds. Essentially in our case, we model the PtrNet in such a way that it takes a candidate point sets (point sets representing the walls) as input and output fixed point set polygons representing rooms of different shapes. This flexible model allows us to learn multiple rooms of different shapes and sizes in an indoor space. Since the effectiveness of PtrNet largely depends on a good spatial order of input point sets, in this paper we propose different variants of input point sets sorting methods before we feed them into a network. Since sorting the input candidate point sets from a random scan is a challenging task and it can affect the performance of the model significantly, we resort to an alternative network that can identify the entire room as a mask, which is our second approach. The Mask R-CNN takes an image as input and a set of binary masks as label/output. Each mask is for an individual room and points to the area where the room is located on that image. In our case, an input is the image showing the walls, and the corresponding label is a set of masks for each of the rooms in that scene.



The contributions of this paper can be summarized as follows.

\begin{itemize}
    \item We develop first comprehensive deep learning based solutions to automatically identify rooms and 2D layouts of an indoor space from a noisy point-clouds data obtained through mobile devices.
    \item We investigate two different paradigms of deep learning: one is geometric learning and the other is image segmentation based learning for learning rooms and layouts of an indoor space, and subsequently propose two methods to identify rooms of different shapes and sizes.
    \item  Extracting rooms and layouts using deep learning is challenging mainly due to scarcity of 3D indoor scene data. To solve this problem, we  introduce a novel approach to generate a synthetic dataset of indoor scenes.
    \item  We conduct extensive experiments on different datasets to evaluate the performance of our proposed approached in identifying different types of rooms that include rectangular, rectilinear, curved walls, non-right angles and also validate performance with real-world methods.
\end{itemize}

The rest of the paper is organized as follows. We discuss related research on indoor mapping and layout generation in Section~\ref{sec:rel}. Then, we describe the methodology in Section~\ref{sec:method} that include two deep learning based models for room and layout generation. After that, we discuss our dataset generation approach for deep learning models and other experimental setup in Section~\ref{sec:exp}. We present our results both on synthetic test data and real-world test data in Section~\ref{sec:eval}. Finally, we conclude the paper with possible future directions in Section~\ref{sec:conclu}.

\section{Related Works}\label{sec:rel}

Indoor mapping and localization has been an interesting research area due to its wide range of real world applications. In this section, we will review some of the related studies in this field.

\textbf{Specialized Sensors for Indoor Mapping.} Vosselman et al. \cite{vosselman2014design} present an indoor mapping system based on three 2D laser scanners. Using laser scanner data, it estimates planes of floor, ceiling and wall simultaneously. Tardos et al. \cite{tardos2002robust} describe a technique to generate feature-based stochastic maps using Polaroid sonar sensors. Their method extracts environmental features, such as straight lines, corners from noisy sensor data.

\textbf{Smart-phone Based Indoor Mapping.} Pintore et al. \cite{pintore2014interactive,pintore2014effective,pintore2016mobile,pintore2017mobile} introduce several methods to build indoor maps using smartphones.  In \cite{pintore2014interactive,pintore2014effective} they present a system to reconstruct indoor scenes using common smartphone features, such as camera and magnetometer. Although their system struggles in curved walls and narrow spaces, this fascinatingly simple system is able to map multi-room indoor scenes in almost all cases. Pradhan et al. \cite{pradhan2018smartphone} use smartphone-based acoustic mapping system to reconstruct indoor walls, but do not extract rooms from it.

\textbf{Recent Point-clouds Based Indoor Mapping.} In 2015, Google's Project Tango developed an application~\cite{tango2017floorplan} for Tango enabled devices to create rough floor plan. This application detects free spaces, furniture and walls. Although the produced indoor maps mostly preserved their original shapes, their outlines are noisy and fragmented.  Murali et al. \cite{muraliindoor} present a system to generate 2D indoor model from existing 3D point-cloud. However, this method assumes all angles between walls to be right angle. Recently, Angladon et al. \cite{angladon2018room} present an real-time approach to produce accurate floor plans of single rooms using Tango Development Kit. Their approach is similar to Project Tango's Floorplanner application \cite{tango2017floorplan}. Although, they have their own binary classifier to differentiate between clutter and wall. Babacan et al. \cite{babacan2017semantic} propose a deep learning approach towards semantic segmentation of indoor point-cloud. They divide 3D scenes into voxels and classify each voxel into classes (e.g. walls, desks, chairs, humans etc.) using 3D Convolutional Neural Network. None of these approaches provide any information about rooms. Jung et al. \cite{jung2017automatic} propose an automatic room detection system from point-clouds. They estimate the height of the scene by detecting the floor and the ceiling of the scene. Then, they avoid background clutters by selecting those points which lie in between the plane of the ceiling and a horizontal plane just below the ceiling. After that, they project those points onto a 2D binary map and close any opening, such as a doorway, between two adjoining rooms. Thus, the 2D binary map is segmented into regions, which are identified as rooms. Armeni et al. \cite{armeni20163d} propose a system that can semantically segment large scale 3D point-cloud of an entire building. As a first step, they divide point-clouds into spaces (e.g. rooms). However, both of these approaches are performed on complete point-clouds of scenes collected by special equipment like LiDAR or laser sensors. Moreover, if two adjoining rooms have large gaps in the shared wall, these methods will detect both of them as a single room.  Recently, A deep learning based approach \cite{liu2018floornet} propose a neural network architecture FloorNet, which turns a RGBD video stream covering a large 3D space into floor-plan. It uses not only point-clouds, but also RGB images. Although FloorNet is able to generalize well from low amount of training data, it fails to estimate layout during noisy point-clouds, missing portions or incomplete scanning.

\section{Methodology}
\label{sec:method}
Our approach consists of two major steps: wall detection from point-clouds and room detection from walls. We propose two deep learning strategies, one that learns geometric borders in terms of sequence of points, and the other that learns masks of different rooms, to identify rooms from detected walls in the first phase. In this section, we explain these steps in detail.

\subsection{Acquisition of Point-clouds}

We acquire 3D reconstruction of indoor scenes in the form of 3D meshes or point-clouds with the help of smart devices like Google's Tango. Usually in 3D meshes or point-clouds of indoor scenes, wall surfaces are occluded by furniture like tables, chairs, beds, wardrobes, shelves etc., and also by humans, curtains and other background clutters. This makes indoor modeling a more challenging task. We use wall detection module to exclude all kinds of background clutters and keep only the wall surfaces. Even after wall extraction, we cannot fully recover shapes of the rooms. Because, often user-generated 3D meshes or point-clouds have some missing portions due to openings like doors and windows or due to an incomplete scan. Thus, often extracted walls have gaps in them. Moreover, the wall detection module cannot always remove all of the background clutters. As a result, extracted walls can often be distorted and noisy. To handle these challenges, we use two deep learning approaches which are trained to extract rooms even when there are gaps and noises in extracted walls. 

\subsection{Wall Detection} \label{sub:wall}

The first step of our process is to identify walls of different rooms from given point-clouds generated from mobile scan of the indoor space. Since the 3D scan may be of different formats like 3D mesh or simple point-clouds, as a first step of the wall detection process, we first need to sample points representing the entire indoor space and then detect walls from these point sets. 


\subsubsection{Sampling Points}

We start with the 3D reconstruction of the indoor scene of interest. Our target is to pass a point-cloud with fixed $N$ number of points to the next stage. If the 3D reconstruction is already captured in the point-cloud format, we just randomly select $N$ points from it to get a sampled point-cloud. If reconstruction is done in triangular mesh format, we do the following to get a sampled point-cloud. A 3D triangular mesh is a collection of triangles. To sample a single point, a triangle is randomly chosen with a weighted probability proportional to its area. The larger the area of a triangle, the more probability it has to get chosen. Then, a point is uniformly sampled within the boundary of the selected triangle. We repeat this process $N$ times to sample $N$ points. 



\subsubsection{2D Histogram}

\begin{figure}[!ht]
    \centering
        \subfloat[2D histogram]{\includegraphics[width=0.48\columnwidth]{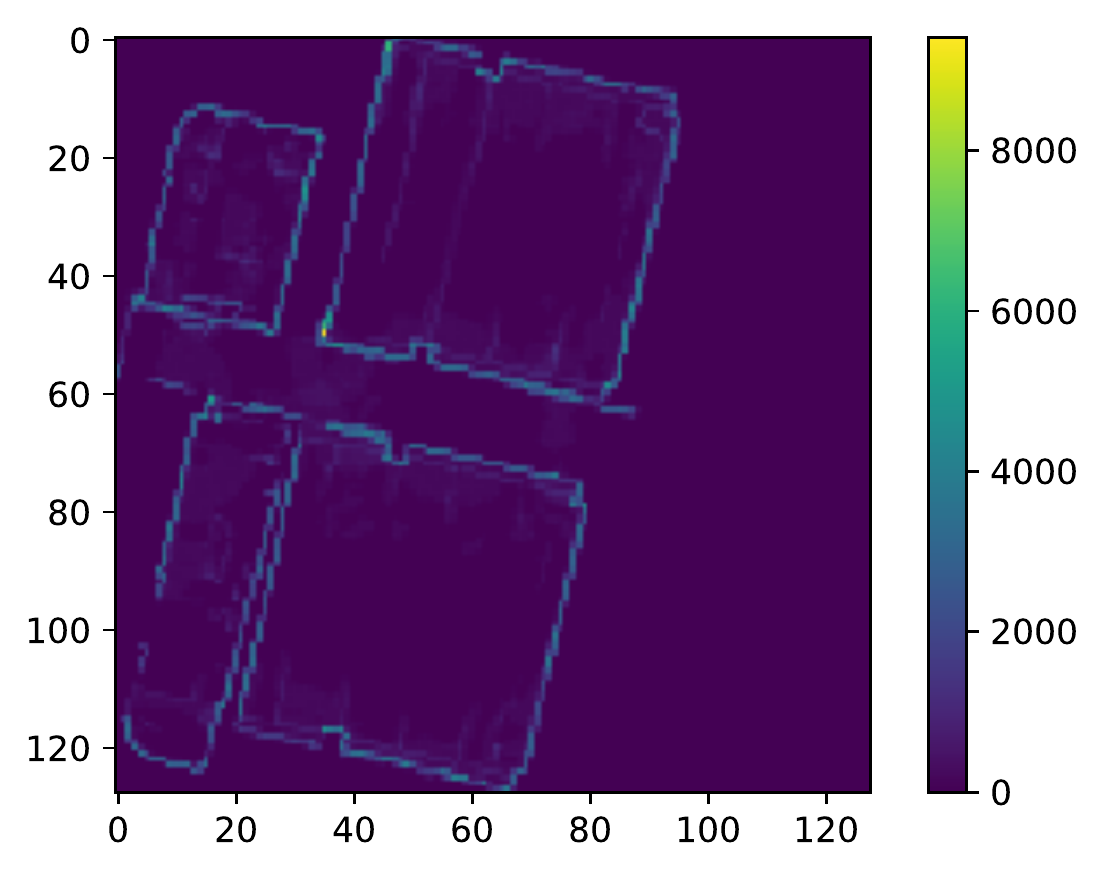}}
        \qquad
        \subfloat[Extracted walls]{\includegraphics[width=0.4\columnwidth]{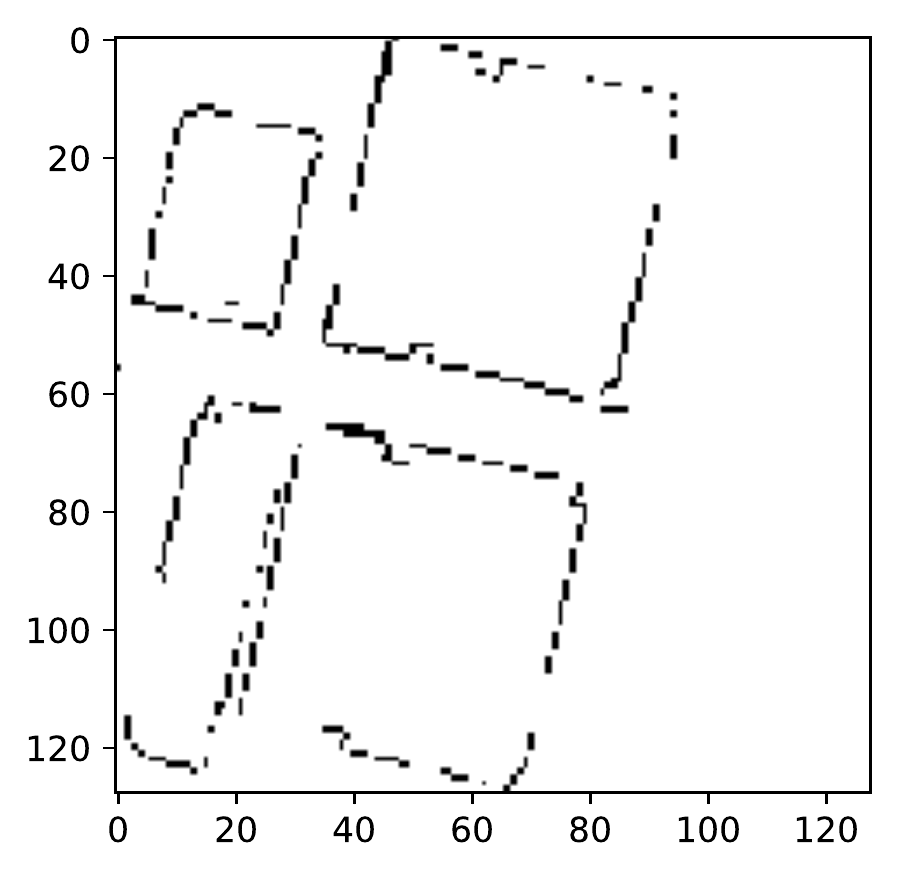}}
    \caption{Wall detection by 2D histogram}
    \label{fig:hist}
\end{figure}

We adopt a 2d histogram based technique, originally proposed by Adan et al. \cite{adan20113d}, to identify only the points representing the walls from the entire point-cloud of size $N$. We first divide the floor into $D \times D$ bins to get a square shaped grid structure. This grid is big enough to fit the $(x, y)$ coordinates of all points within it. Then, we project all points to the floor and count the number of points that fall into each bin. This gives us a 2D histogram like Figure \ref{fig:hist}(a). A bin tends to have more points if there is a wall on it. Here, we call these bins ``wall bins''. The shape of the walls is clearly visible is Figure \ref{fig:hist}(a). 

We extract wall bins by a threshold mechanism. A bin having points less than the threshold is not considered a wall bin. Using a fixed threshold here does not yield a desired result in our case. Hence, we find the maximum number of points $n_{max}$ in a bin among all bins and use $n_{max} / 4$ as the threshold value by empirically evaluating from a large number of scans of real rooms. Finally, we get a $(D, D)$ shaped image (pixels or points) showing the walls like in Figure \ref{fig:hist}(b). Later, we use this image to extract rooms by using deep learning techniques.

\subsection{Room Detection using Pointer Network}

In our first deep learning approach, we resort to learning the geometry of different rooms as a sequence of points. Thus, we use the geometric learning framework, namely Pointer Network (in short, PtrNet) to identify locations and layouts the rooms from extracted walls of our previous step. Next, we first give an overview of PtrNet and then explain how we adopt PtrNet in identifying rooms as a sequence of points.

\subsubsection{PtrNet}

A Pointer Network is a seq2seq (sequence-to-sequence) architecture with attention mechanism, where the output is a sequence of input indices of input vectors. That means, PtrNet can select only those vectors as output that are already present in input sequence. PtrNet supports variable-length inputs. Mathematically, let $P = \{P_1, \ldots, P_n\}$ be an input sequence of vectors, $C = \{C_1, \ldots, C_m\}$ an output sequence $C = \{C_1, \ldots, C_m\}$ , where $C_i \in \mathbb{N}$ and $1 \leq C_i \leq n$ for all $C_i \in C$. Here, $\theta$ is RNN parameters. Then the probability of output sequence $C$ is computed as follows: 
$$ p(C|P; \theta) = \prod_{i=1}^{m} p(C_i|C_1,\ldots,C_{i-1},P;\theta). $$
This model will essentially try to estimate $\theta^{\ast}$ by maximizing probabilities, as stated below,
$$ \theta^{\ast} = {argmax}_{\theta} \sum_{P,C} \log{p(C|P;\theta)}. $$ 
To find $p(C_i|C_1,\ldots,C_{i-1},P;\theta)$, a typical seq2seq model uses softmax over fixed sized output dictionary. As a result, it cannot be applied to problems where output dictionary size is equal to input length. PtrNet uses attention mechanism to solve this problem, which can be stated as follows:

\begin{align*}
    u^i_j &= v^T \tanh(W_1e_j+W_2d_i) \quad j \in (1,\ldots,n) \\ 
    p(C_i|C_1,\ldots,C_{i-1},P) &= \text{softmax}(u^i)
\end{align*}

Softmax over vector $u^i$ of length $n$ is applied to produce the output sequence. Similar to the attention mechanism, $v, W_1, W_2$ are learnable parameters in this case. PtrNet does not use encoder states $e_j$ to provide information to decoder. Rather, it uses $u^i_j$ to point back to input sequence. So, in short, PtrNet can learn a model where input is a sequence of vectors, output is discrete and output can be constructed with input vectors.

\subsubsection{PtrNet in Room Identification}
To solve our room detection problem using PtrNet, we formulate the problem such that it fits into the structure of PtrNet framework. After the wall extraction phase, we have two types of pixels - those who have walls on them and those who do not have walls on them. We take the $(row, column)$ of wall pixels and treat them as coordinates of points. Thus, we have a point set $S'$. Then, we shift and scale the coordinates of points in $S'$ to be within a $[-1, 1]\times[-1, 1]$ square. Then, we sample a fixed number of points $P_n$ from $S'$ to get a new set $S$.  This point set $S$ is the input of PtrNet. Our  target is to predict the borders of rooms. For this, PtrNet selects points from input points to create new sets of points. Each set is a polygon with fixed $b$ number of points. Each polygon marks the border of a single room. In other words, PtrNet will only say which input points are border points.

\begin{figure}[!ht]
\centering
\includegraphics[width=0.8\columnwidth]{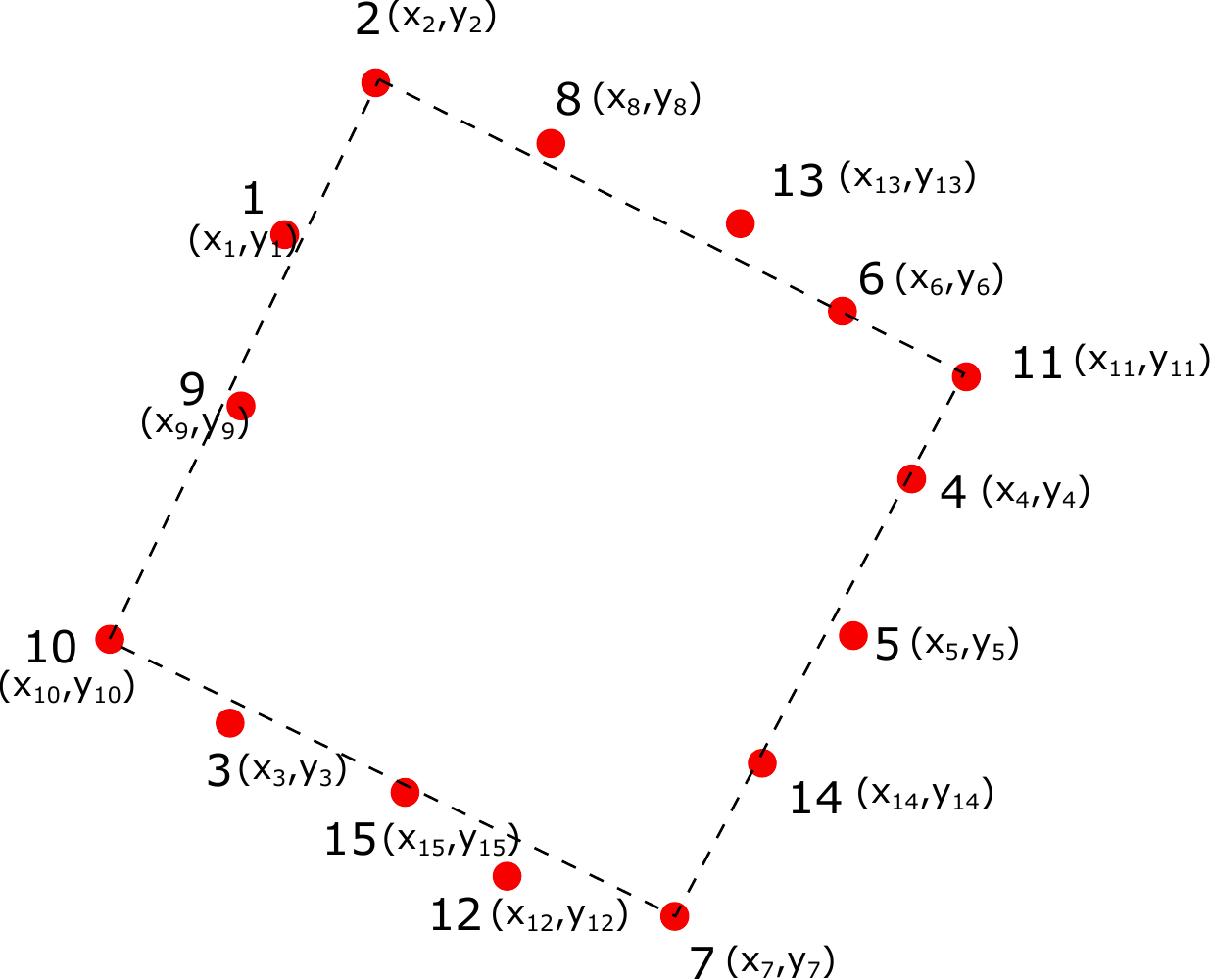}
\caption{Sample points representing the walls of a single room.}
\label{fig:ptrnet-sample}
\end{figure}

Figure \ref{fig:ptrnet-sample} shows a toy example that illustrates the above concept. For ease of demonstration, we consider a small set of 15 points to represent the detected walls of a single room. The room outline is drawn with dashed line. Also, let us suppose, each room has 4 border points. Then, point set $\langle 7, 11, 2, 10 \rangle$ creates a polygon that marks the room border, which will be the output of our PtrNet framework.

\begin{figure}[!ht]
\centering
\includegraphics[width=\columnwidth]{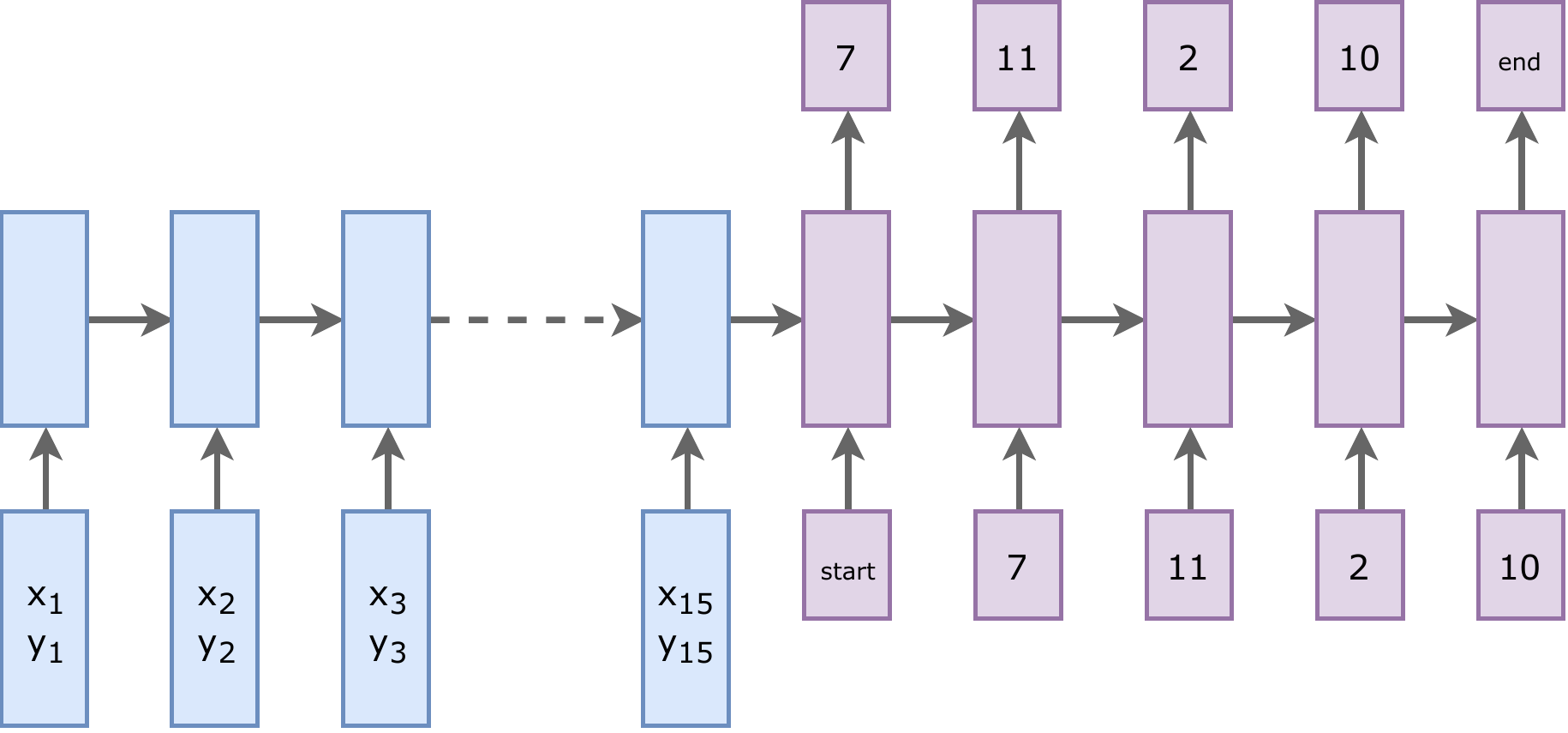}
\caption{Unrolled encoder decoder of PtrNet. For sample scenario.}
\label{fig:ptrnet}
\end{figure}

The PtrNet framework using this toy example is shown in Figure \ref{fig:ptrnet}. We first input all 15 points in encoder network. For the purpose of demonstration, RNN cell is unrolled for each time step. Then the decoder network outputs the indexes of border points.

We extend the above concept to identify multiple rooms. Let us assume that output sequence $C$ consists of $K$ rooms, and each rooms have $b$ border points. Then the length of the output sequence would be $b \times K$. Here, `index' of border points of $1^{st}$ room is $\langle C_1, C_2, \ldots , C_b \rangle$. Similarly, `index' of border points of $k^{th}$ room, where $1 \leq k \leq K$, is $\langle C_{bk-b+1}, C_{bk-b+2}, \ldots , C_{bk} \rangle$.

\subsubsection{Input Ordering}
PtrNet treats a input set of points $S$ as a sequence, not as a set. After we extract walls from point-cloud, we randomly sample a fixed $P_n$ number of points from it. So, after sampling, points are in random order. According to Vinyals et al.~\cite{vinyals2015order}, order of points in input sequence has an effect on PtrNet's performance. So, one of the main challenges is to order points in $S$ before we feed them to PtrNet. Here, we propose three schemes to order the elements in $S$.

\textbf{PtrNet-Random:} Since we randomly pick points from walls, the input sequence to the PtrNet can be truly random, i.e. the input set $S$ is in random order. In this approach, we feed these random set of points (without considering any spatial order), to PtrNet, and evaluate the performance of this PtrNet-Random architecture in detecting rooms.

\textbf{PtrNet-TrueSort:} In this case, we manually sort the points in the desired order and feed into the PtrNet architecture, namely PtrNet-TrueSort. First, points from each room in a sample is sorted in anti-clockwise order. So, each room has a sorted point list. These lists are appended to form input sequence. We refer to this ordering as `true-sort'. This does not represent real world situation, as we cannot know beforehand which point belongs which rooms. It can only be done if we generate our own dataset synthetically. However, the main purpose of PtrNet-TrueSort is to simply show how order affects the performance of PtrNet in our problem.

\textbf{PtrNet-PseudoSort:} Since no known approach can do the true sort of the input point sets, we develop a heuristic to  sort the points, which approximate the true sorting of the points. Here, first, input set $S$ is randomly picked from the given set of points (similar to the procedure described in PtrNet-Random). Then we order the points in $S$ based on proximity of last selected point. Algorithm \ref{alg:pseudo} shows the process of our proposed ordering. First we take an list $L$ and insert the first element of $S$ into it. We also remove the first element from $S$. Then, we enter into a loop. To select which point to remove, we take the last inserted element $q$ in $L$. Then, we find the closest point $p \in S$ to $q$. Then we remove $p$ from $S$ and insert it into $L$. We keep selecting new $p$ until $S$ is empty. Finally, $L$ gives us the ordering. We refer the PtrNet based on this type of ordering as PtrNet-PseudoSort.


\begin{algorithm}
\DontPrintSemicolon
\caption{Pseudo-Sort Algorithm}
\label{alg:pseudo}
\SetKwProg{Pseudosort}{Function \emph{Pseudosort}}{}{end}
\SetKwInOut{Input}{input}
\SetKwInOut{Output}{output}

\Input{Shuffled input sequence $S$}
\Output{Ordered input sequence $L$}
\Pseudosort{$S$}{
    $L \gets \{S_1\}$\Comment*[r]{Taking the first element of $S$}\
    $S \gets S - \{S_1\}$\;
    \While{$S \neq \{\}$}{
        Find point $p \in S$ that is closest to the last element in $L$\;
        $L \gets L \cup \{p\}$\;
        $S \gets S - \{p\}$\;
    }
    \textbf{return} $L$\;
}
\end{algorithm}

After conducting experiments, in evaluation section (Section \ref{sec:eval}), we show how PtrNet performance varies depending on which ordering method we choose

\subsection{Room Detection using Mask R-CNN}

As our second deep learning strategy, we adapt a deep learning framework, Mask R-CNN, commonly used for image object segmentation, to identify locations and layouts of the rooms from extracted walls. For this, we train an image segmentation neural network to segment rooms from the image of extracted walls. We first discuss the basics of Mask R-CNN and the we discuss how we use this approach to our purpose.

\subsubsection{Image Segmentation Network: Mask R-CNN}

Mask R-CNN \cite{he2017mask} is an object segmentation Convolutional Neural Network (CNN), that can identify each pixel of an object from an image. As explained in Section \ref{sub:wall}, we can obtain the image of the top view of walls from a point-cloud. After that, we identify the rooms from the extracted walls. As an instance of segmentation network, Mask R-CNN suits perfectly for the task of segmenting individual rooms from that image. 

A typical object detection model needs someone to annotate the objects in an image using bounding boxes. However, for training Mask R-CNN, a sample is an image and some binary masks. An image can have several masks, each mask marks the pixel-wise location of an object. Mask-RCNN has two steps. First, it produces region proposals on where an object might be. Second, a classifier identifies the class of that object and improves on bounding box and produces a pixel level mask of the object. 

In the first step, to get a set of Regions of Interest (RoI), Mask-RCNN uses RoI Align. RoI Align network produces multiple bounding boxes that may contain an object. These boxes are merged and refined to one bounding box using a regression model. In the second step, it does semantic segmentation to every bounding box. The first step guarantees that each bounding box will have at most one class. So, this is like a pixel level binary classifier. In the bounding box, a pixel is 1 if it is a part of that object, 0 otherwise. 



\subsubsection{Mask R-CNN in Room Segmentation}

\begin{figure}[!ht]
    \centering
        \subfloat[Extracted wall]{\includegraphics[width=0.35\columnwidth]{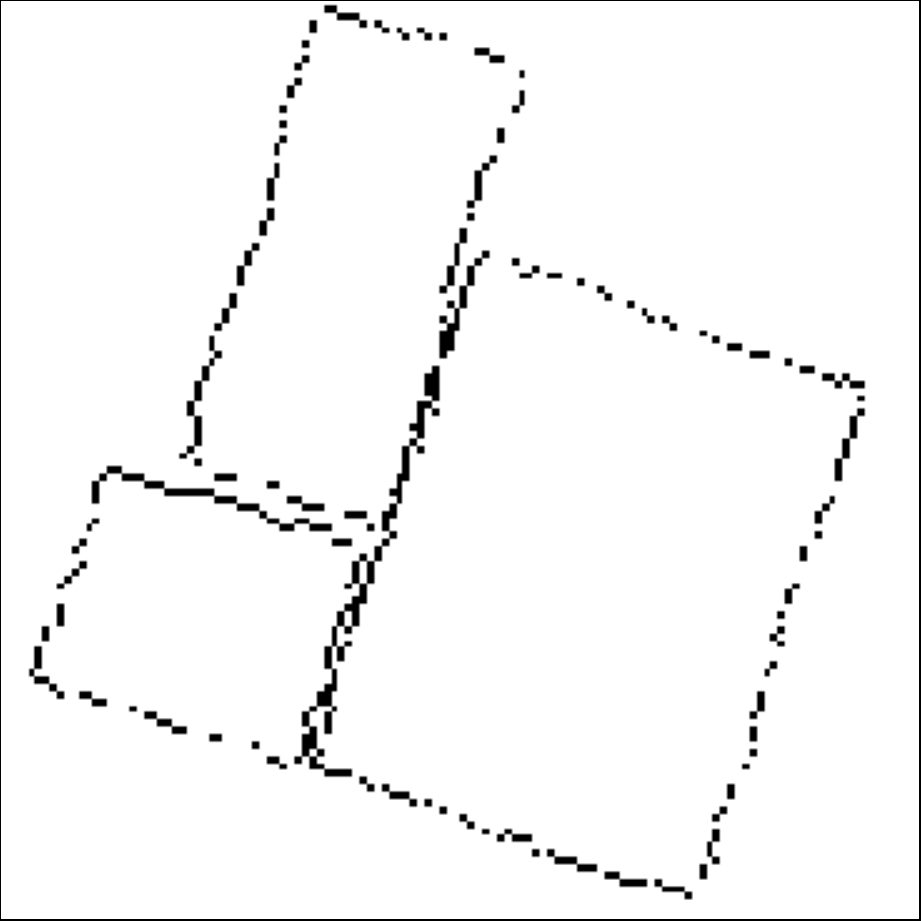}}
        \qquad
        \subfloat[Room mask 1]{\includegraphics[width=0.35\columnwidth]{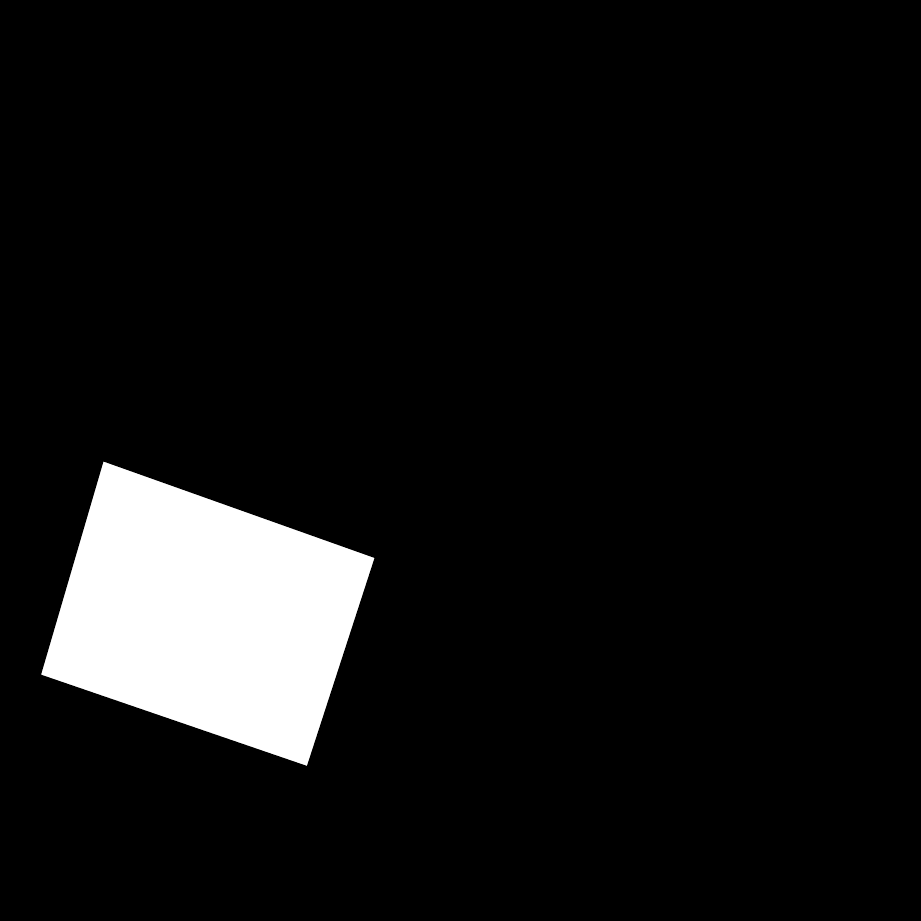}}
        \\
        \subfloat[Room mask 2]{\includegraphics[width=0.35\columnwidth]{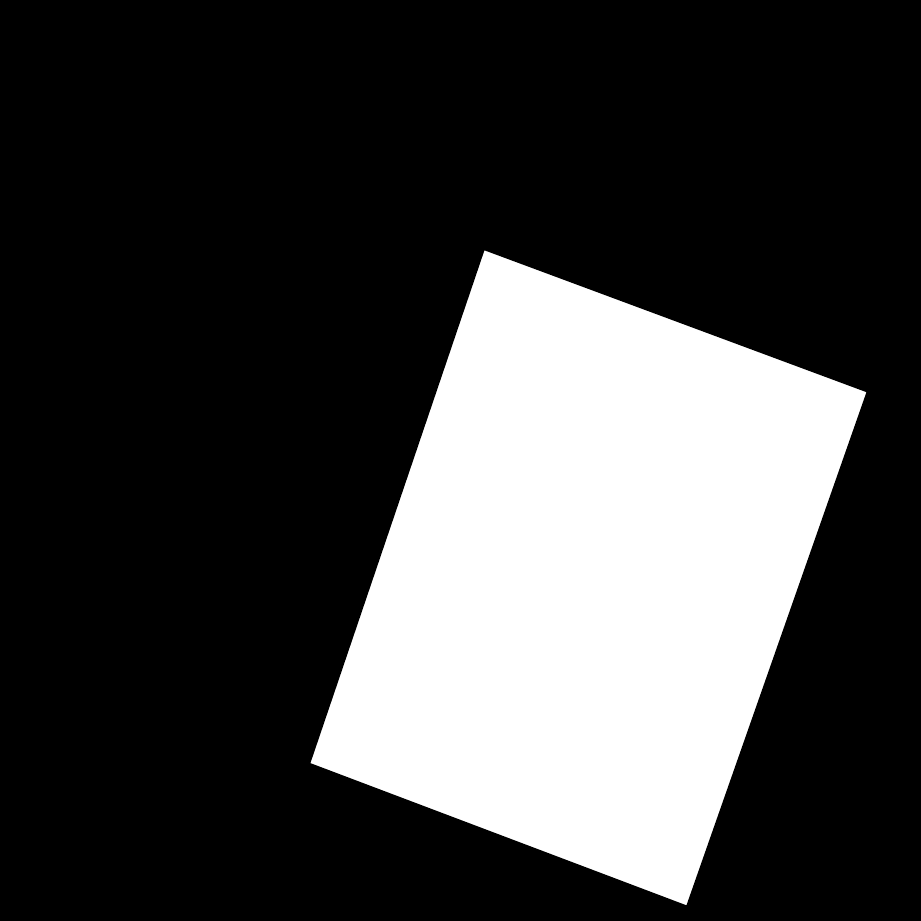}}
        \qquad
        \subfloat[Room mask 3]{\includegraphics[width=0.35\columnwidth]{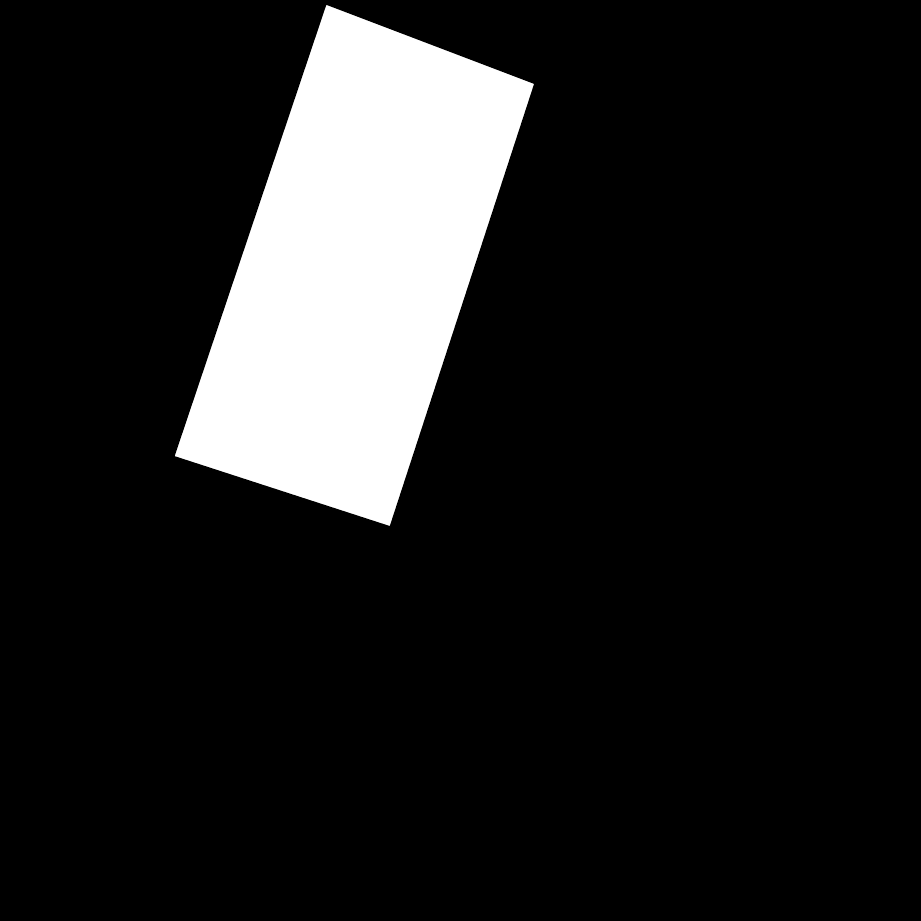}}
    \caption{A sample scenario for Mask R-CNN: 3 rooms and masks for each of them}
    \label{fig:mrcnn-room}
\end{figure}

Mask R-CNN takes an image as input and a set of binary masks as label/output. Each mask is for an individual object and points to the area where the object is located on that image. For room segmentation task using Mask R-CNN, an input is the image showing the walls (Figure \ref{fig:mrcnn-room}(a)), and the corresponding label is a set of masks for each of the rooms in that scene (Figure \ref{fig:mrcnn-room}(b-d)). In the training phase, we provide a dataset consisting both input images and corresponding masks. Mask R-CNN learns from that dataset. After learning, Mask R-CNN takes an image as input and spits out masks as output. Each predicted mask shows the location of a room.

\begin{figure}[!ht]
    \centering
        \subfloat[Predicted masks]{\includegraphics[width=0.4\columnwidth]{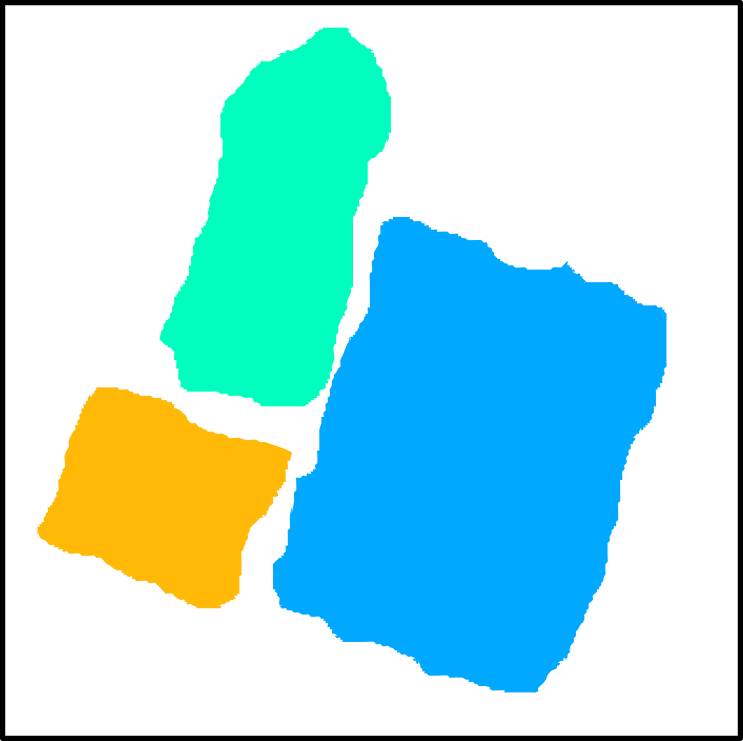}}
        \qquad
        \subfloat[Mask border]{\includegraphics[width=0.4\columnwidth]{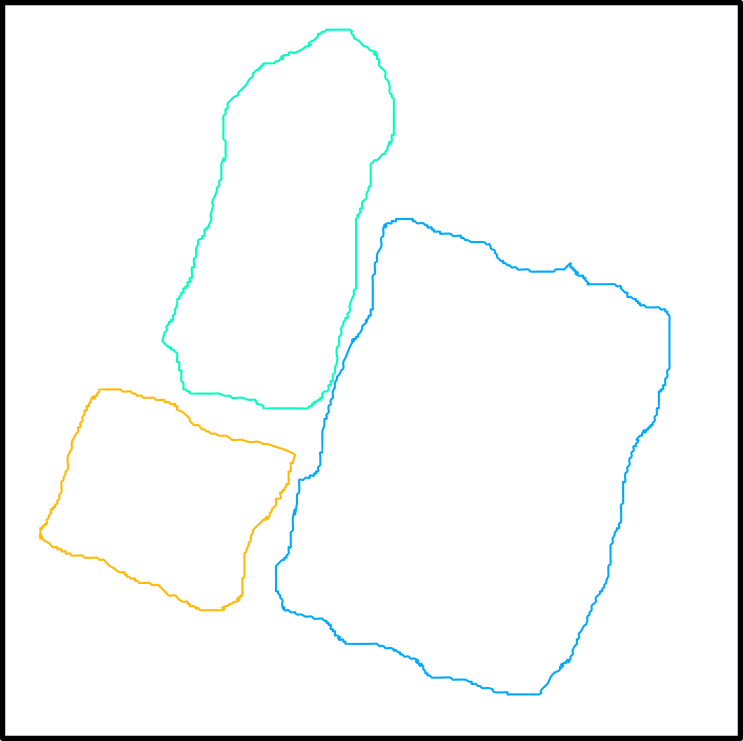}}
    \caption{Extracting borders from the masks of predicted rooms}
    \label{fig:mrcnn-pred}
\end{figure}

Finally for each mask, we have to extract the border of the corresponding room. For this, we obtain the borders of the masks. Given a binary mask, we apply \emph{contour detection algorithm} to find the border pixels of the masked region. A contour is a closed curve connecting all the points along the boundary of an object having same color. After detecting contours, we get a set of pixels for each of the binary masks, that marks the border of the corresponding room as shown in Figure \ref{fig:mrcnn-pred}.

\section{Experimental Setup}\label{sec:exp}

For detecting rooms from extracted walls like Figure \ref{fig:hist}(b), we take a learning based approach. Any deep learning approach requires training data to learn from. Unfortunately, there is no dataset available. We develop a novel algorithm to generate a large number of indoor scenes with only walls. We train PtrNet and Mask R-CNN on it and test on real indoor scenes.

\subsection{Dataset Generation}
\label{subsec:dataset}
\begin{figure}[!ht]
    \centering
        \subfloat[]{\includegraphics[width=0.45\columnwidth]{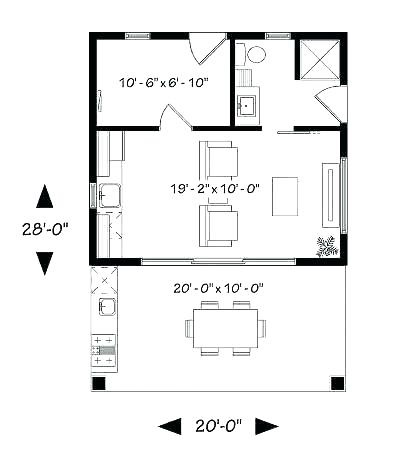}}
        \quad
        \subfloat[]{\includegraphics[width=0.45\columnwidth]{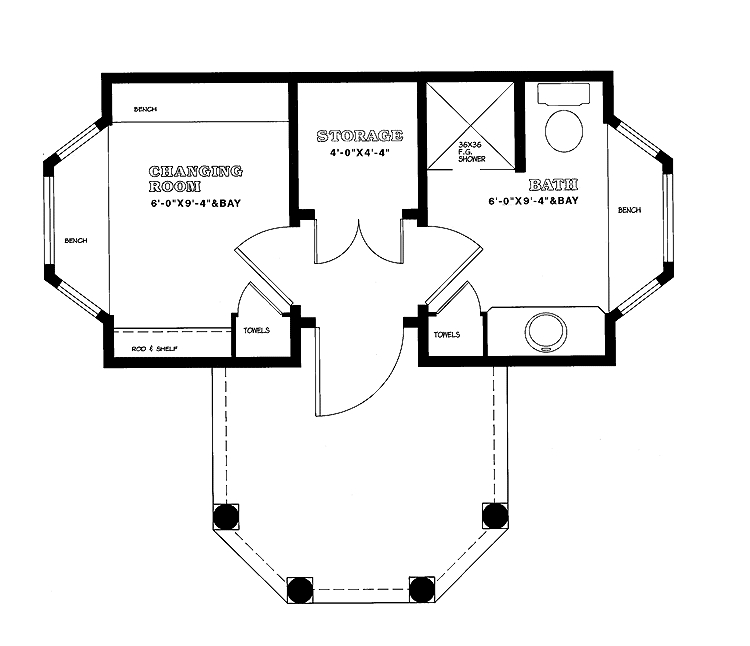}}
    \caption{Examples of real indoor layout}
    \label{fig:real-floor}
\end{figure}

To generate synthetic indoor layout, we first analyze a set of real indoor layouts like Figure \ref{fig:real-floor}. Walls in these layouts include shapes like, rectangular, rectilinear, curved walls and non-right angles etc. In an attempt to make our dataset realistic, we generate these shapes.

Two networks expect input data in two different formations - PtrNet as sequences, Mask R-CNN as images. Generating a sample for both networks, share some common tasks. First we explain the common portion, then we describe how we process them differently.

\begin{figure}[!ht]
    \centering
    \includegraphics[width=\columnwidth]{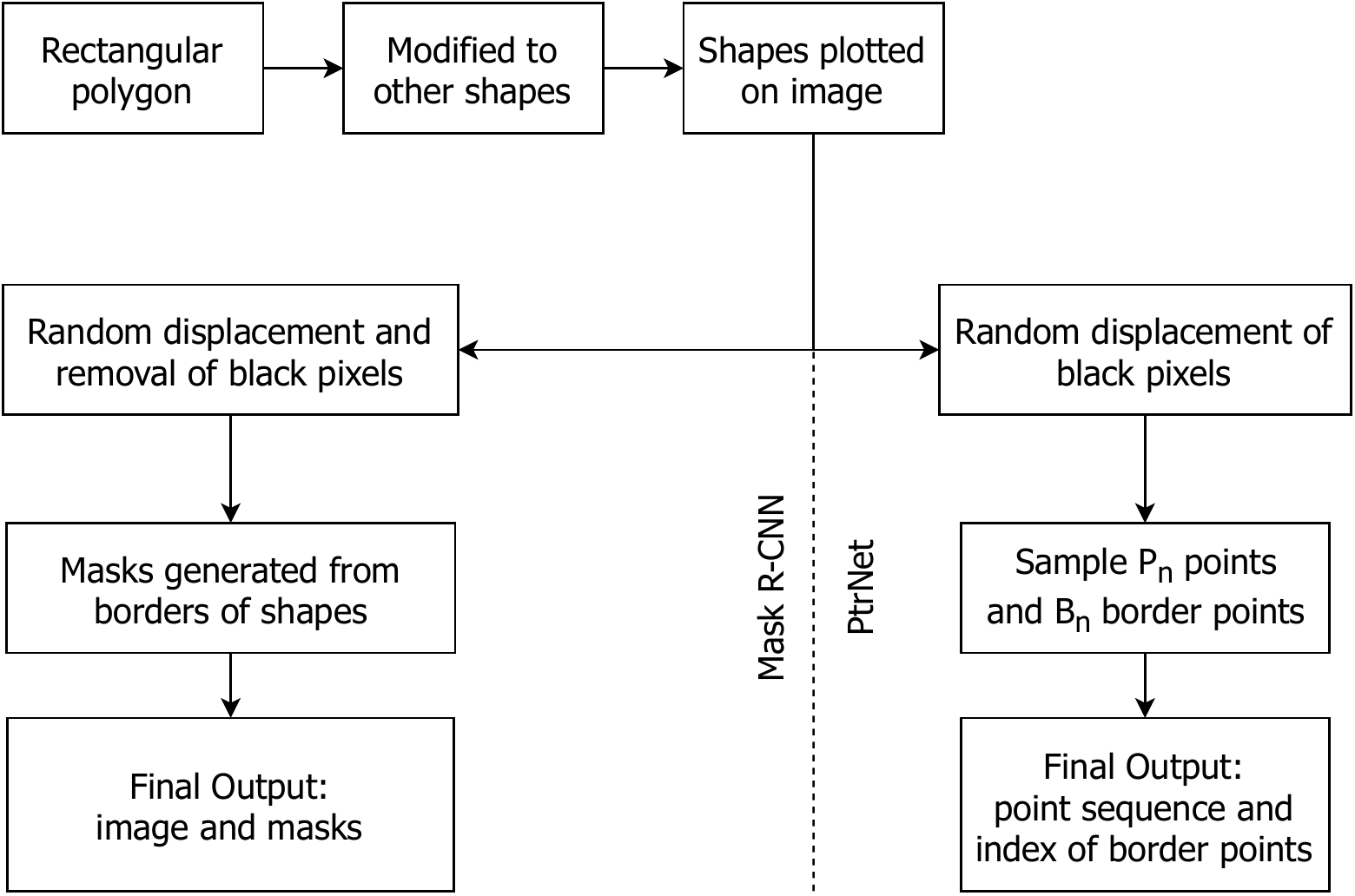}
    \caption{Workflow of data generation}
    \label{fig:data-gen-flow}
\end{figure}

\subsubsection{Rectangular Polygons}

Our approach is to first generate rooms in rectangular shapes and modify them later to create other shapes. So, first we will show how we generate a single rectangle shaped room. Our process requires a starting point $P$ and an unit alignment vector $\bm{v}$. The alignment vector determines the orientation of the rectangle. We uniform randomly select length $l$ and width $w$ of the rectangle within interval $(l_{max}, l_{min})$. Then, we calculate the four corners of the rectangle: $P_1 = P$, $P_2 = P_1 + \bm{v}l$, $P_3 = P_2 + \bm{v_{\perp}}w$, $P_4 = P_3 - \bm{v}l$, where $\bm{v_{\perp}}$ is 90$^\circ$ anticlockwise rotation of $\bm{v}$. 


\begin{figure}[!ht]
    \centering
    \includegraphics[width=0.6\columnwidth]{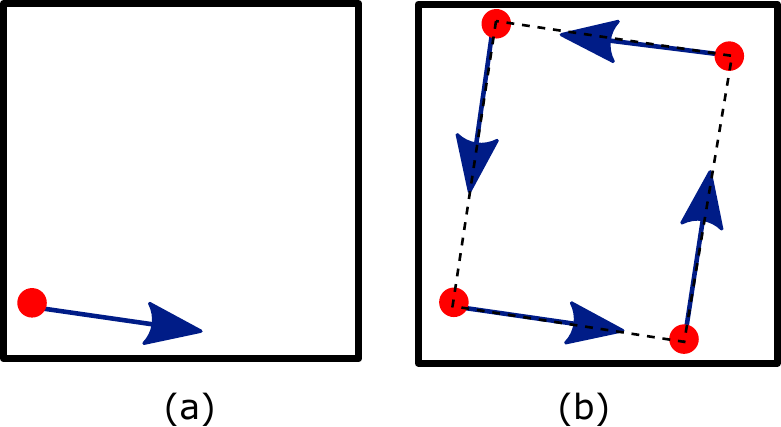}
    \caption{Generation of a rectangular room. Left: a point and an alignment vector. Right: corner points.}
    \label{fig:room-gen}
\end{figure}

\begin{figure}[!ht]
    \centering
        \subfloat[Single rectangle]{\includegraphics[width=0.3\columnwidth]{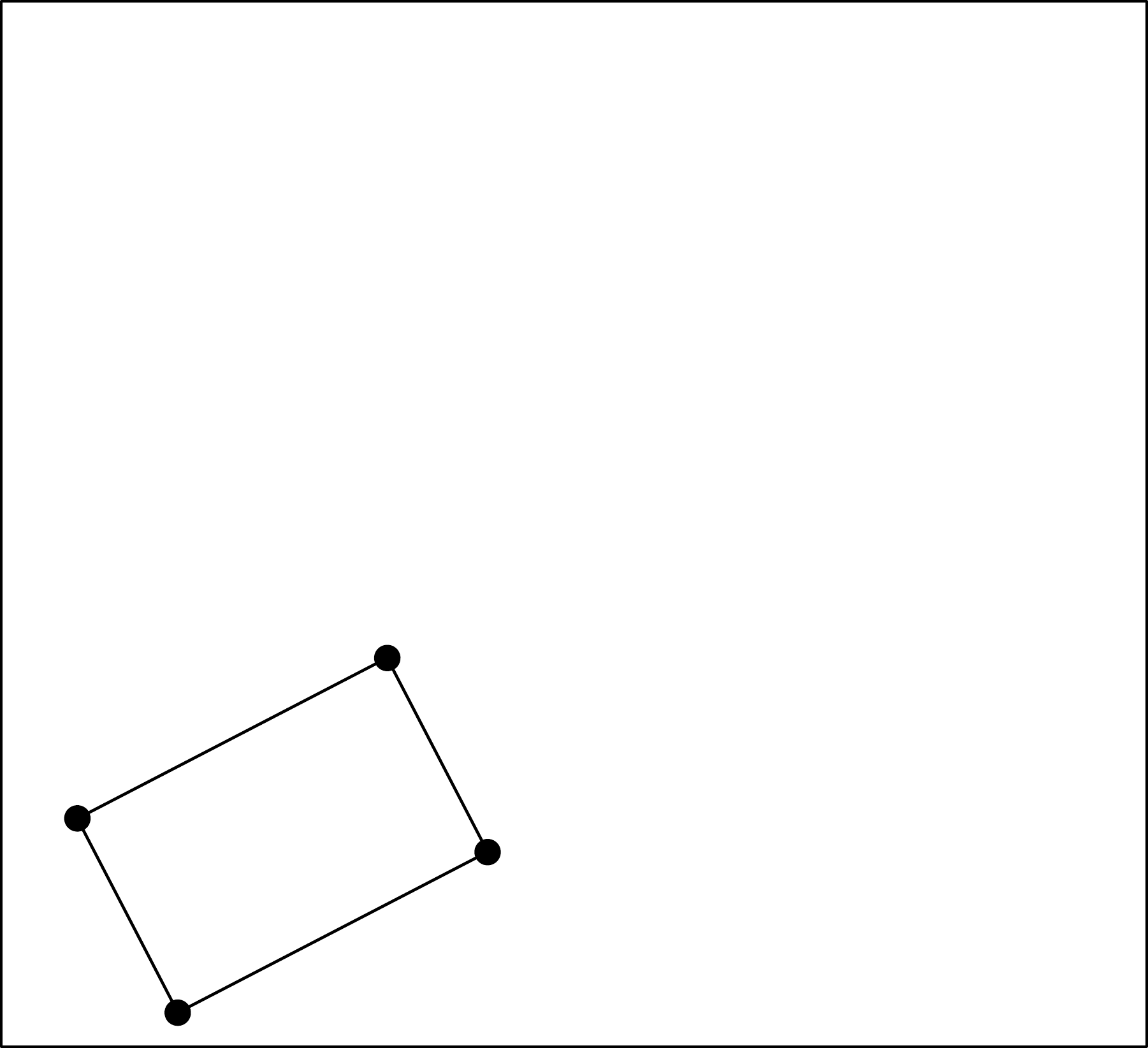}}
        \quad
        \subfloat[Multiple rectangles]{\includegraphics[width=0.3\columnwidth]{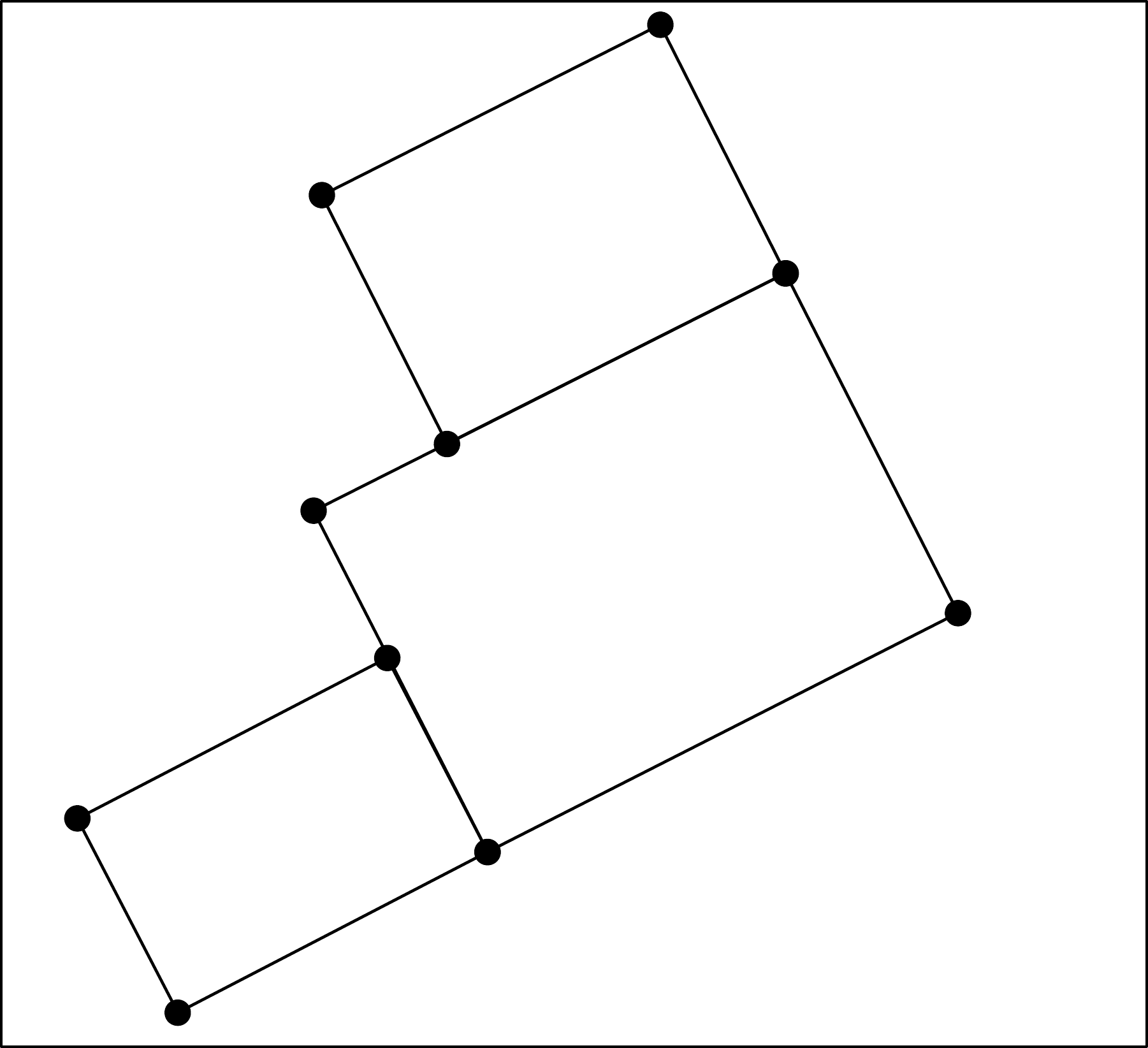}}
        \quad
        \subfloat[Slightly shrunken Rectangles]{\includegraphics[width=0.3\columnwidth]{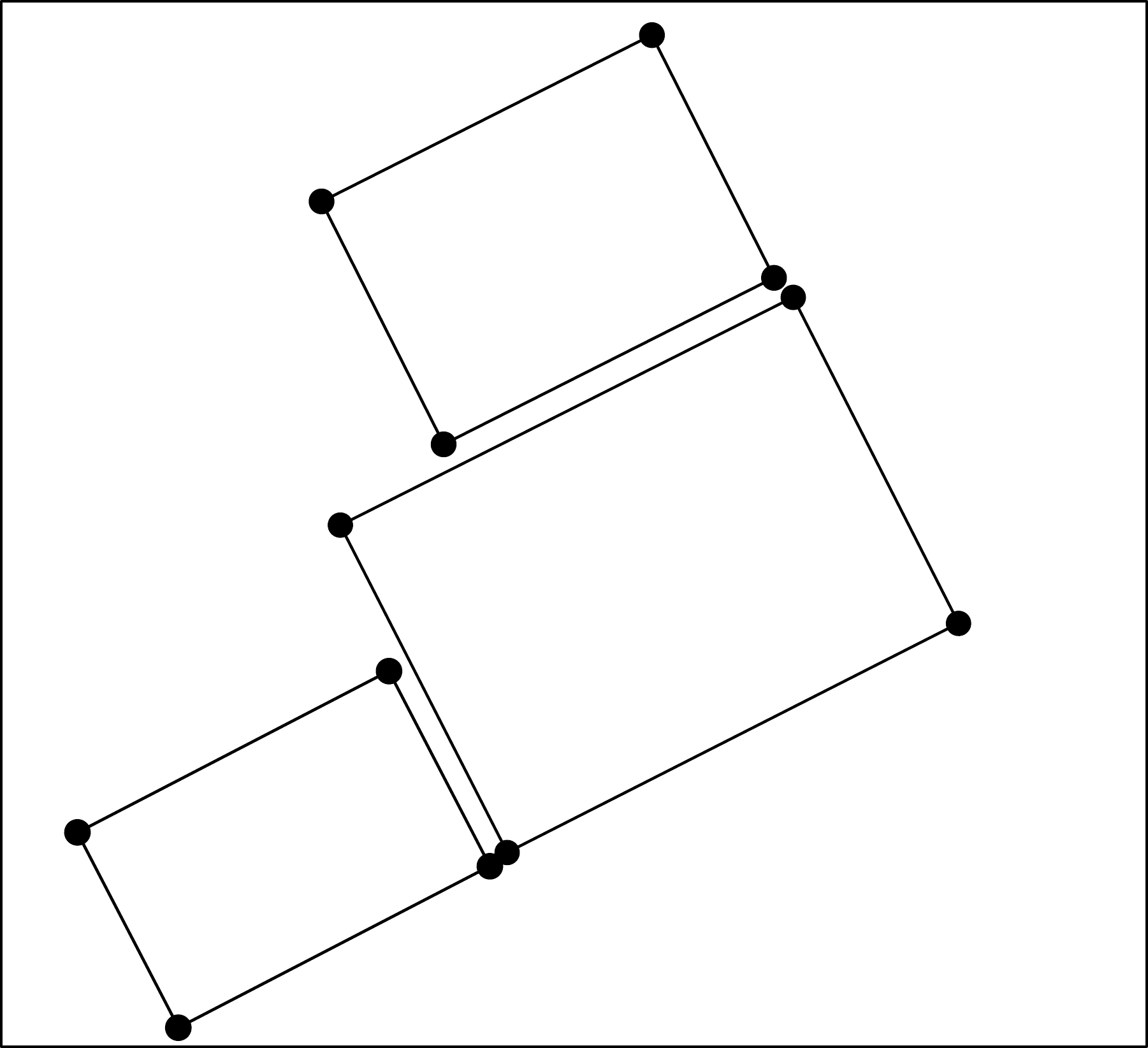}}
    \caption{A sample data point generation process}
    \label{fig:sample}
\end{figure}

To generate a second rectangle, we use one of the corner points of the first rectangle as new starting point. So, the second rectangle is just beside the first rectangle. We again randomly select length $l$ and width $w$ of the rectangle. Similarly, to generate more rectangles, we randomly choose one of the corner points of existing rectangles. We also make sure no new rectangle overlaps any existing one. When we collect point-cloud of adjacent rooms, we get two vertical planes on either side of the shared wall. To incorporate this effect, we shrink each rectangle by a small amount (Figure \ref{fig:sample}(c)).

\subsubsection{Other Shapes}

\begin{figure}[!ht]
    \centering
    \includegraphics[width=\columnwidth]{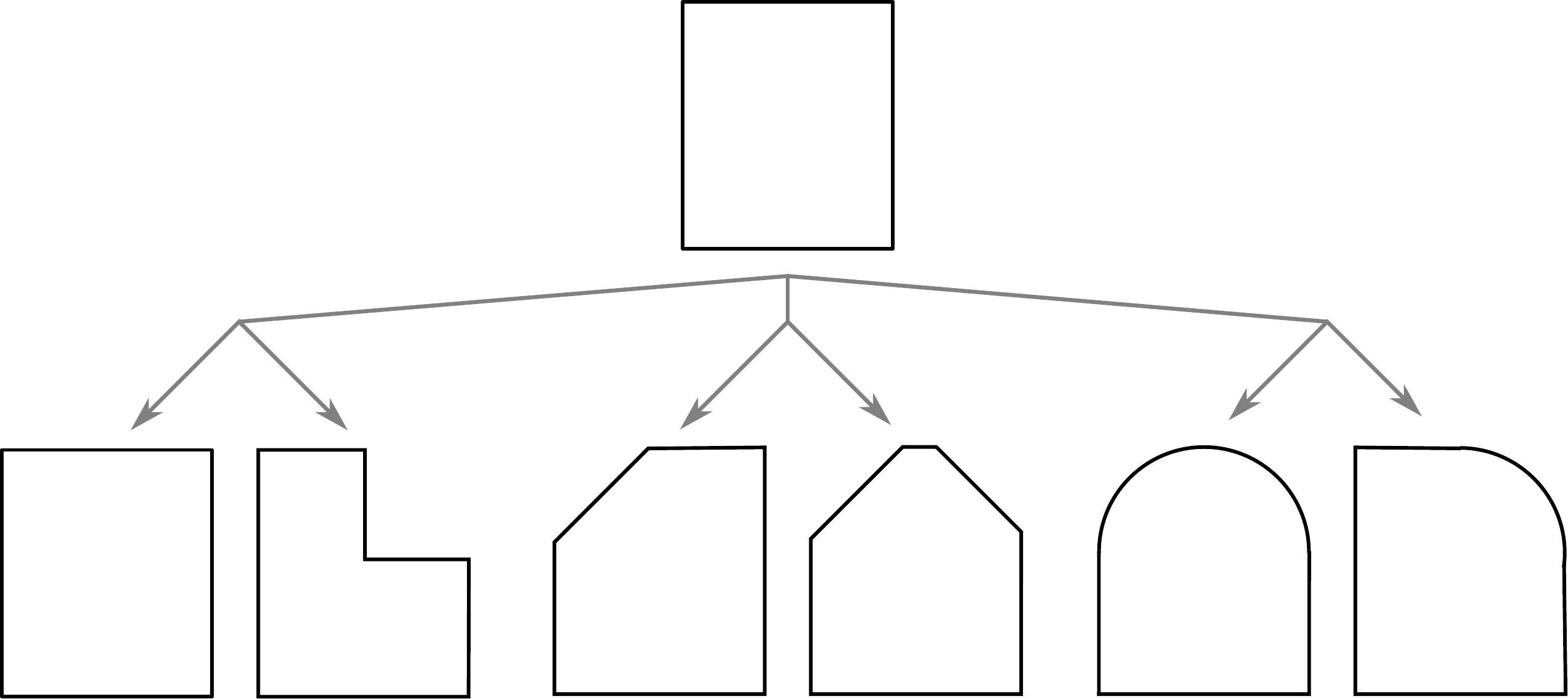}
    \caption{Other shapes created from rectangle. Left to Right: rectangular, rectilinear, two non-right angle shapes, half circular and quarter circular}
    \label{fig:other-shapes}
\end{figure}

We create 3 types shapes from rectangles as shown in Figure \ref{fig:other-shapes} - 
\begin{enumerate}
    \item Shapes with right angle (rectangular and rectilinear)
    \item Shapes with non-right angle 
    \item Shapes with curves (half circular and quarter circular)
\end{enumerate}

\textbf{Rectilinear:} We randomly choose a corner, thus we get two adjacent edges incident from that corner. We replace the corner with three other points - midpoint of first edge, center of rectangle and midpoint of second edge.

\textbf{Shapes with Non-right Angles:} We again randomly choose a corner and thus we get two adjacent edges. We replace the corner with two other points from those two edges. We can choose a different corner and perform same procedure there.

\textbf{Shapes with Curves:} For a half circular shape, we randomly choose an edge and replace it with an half circle with a radius equal to half the length of that edge. For a quarter circular shape, we randomly choose a corner and replace it with an quarter circle with a radius equal to half pf the length of the shortest edge of the two incident edges there.

Now we have a set of shapes. We plot them in black on a $128 \times 128$ sized white image. After that we process differently for two different approaches.

\subsubsection{Data Generation for PtrNet}

Point-clouds collected from Google Tango Devices are usually noisy. And extracted walls like Figure \ref{fig:hist}(b) are not smooth due to clutters. To incorporate this effect, we shift black pixels to neighboring pixels. Then we find the list of $(row, column)$ of black pixels. We treat each black pixels as a point and $(row, column)$ of that black pixel as $(x, y)$ coordinates of that point. Then we compress the points within a $[-1, 1] \times [-1, 1]$ square. Then, we sample at most $P_n$ points from it. This introduces the effect of gaps and incompleteness in our scans. We try four values of $P_n$: 100, 200, 300 and 400. We also assign $10$ border points for each shape. Finally, we have a sequence of points and a sequence of index of border points. To evaluate the effect of ordering on PtrNet's performance, we further apply sorting schemes - PtrNet-TrueSort, PtrNet-Random and PtrNet-PseudoSort. Figure \ref{fig:sample-data}(a) shows a generated sample for PtrNet. 

\subsubsection{Data Generation for Mask R-CNN}

\begin{figure}[!ht]
    \centering
        \subfloat[For PtrNet]{\includegraphics[width=0.45\columnwidth]{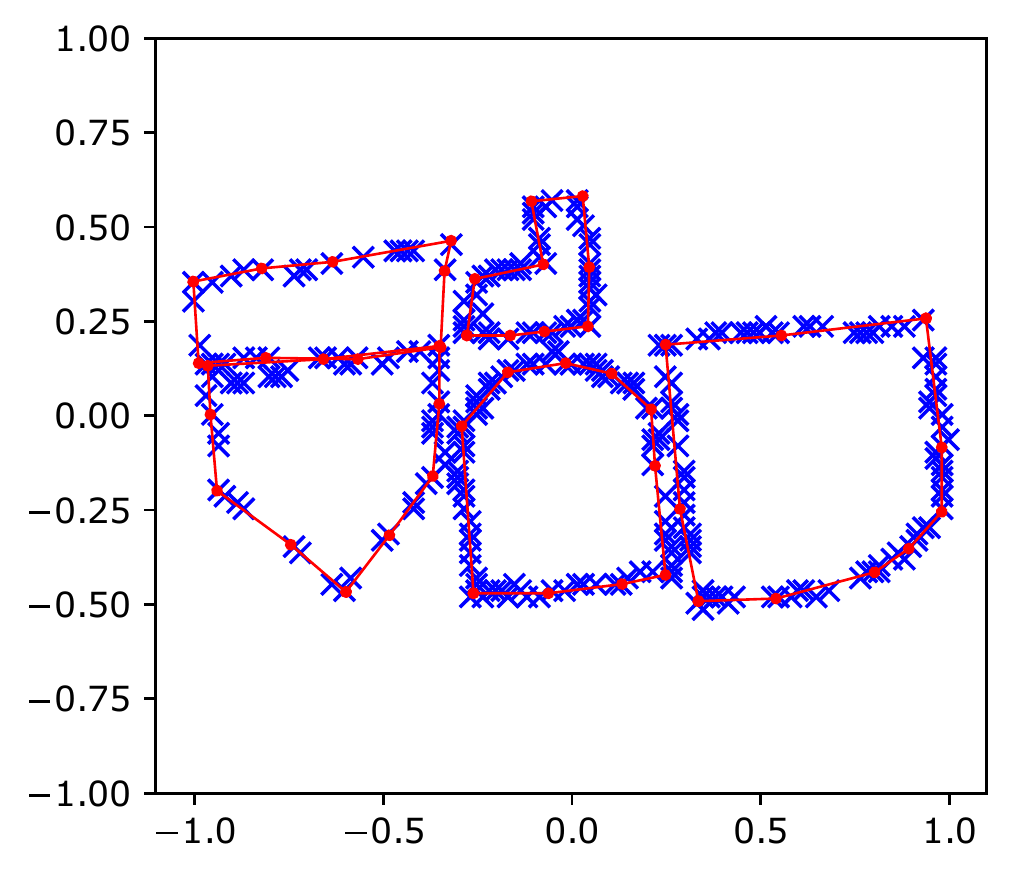}}
        \quad
        \subfloat[For Mask R-CNN]{\includegraphics[width=0.45\columnwidth]{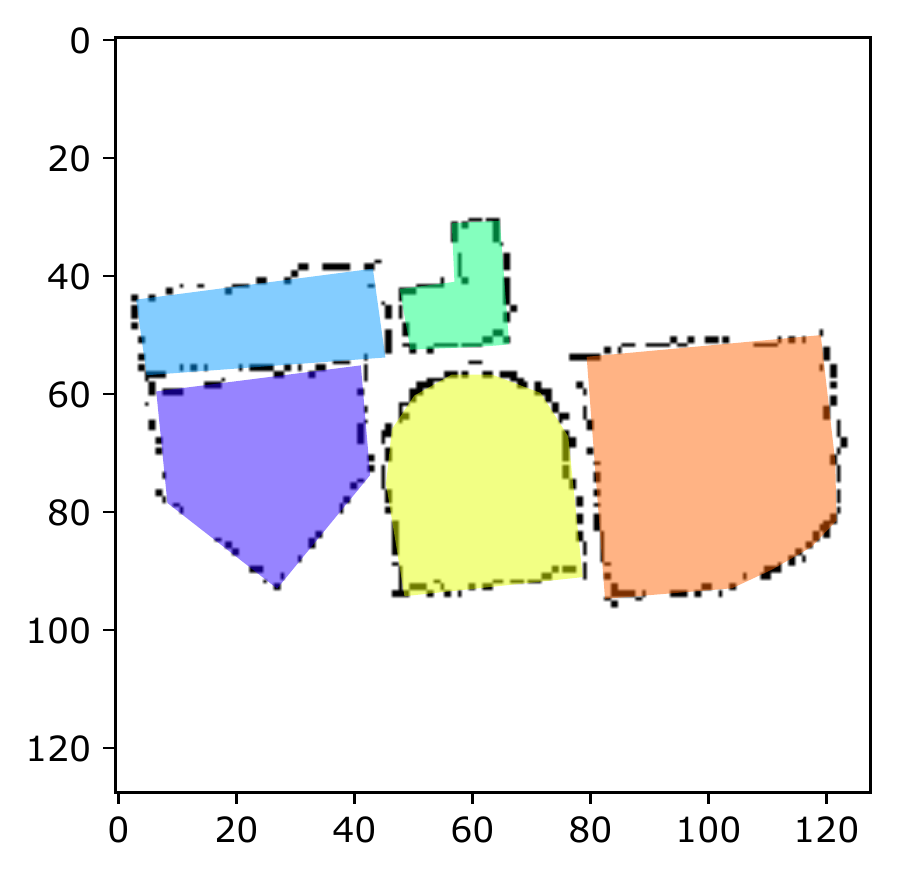}}
    \caption{Generated samples for both approaches. Left: Input points are marked with blue crosses, borders are marked with red dots and lines. Right: Rooms and corresponding masks.}
    \label{fig:sample-data}
\end{figure}

We displace black pixels for this too. We also randomly remove black pixels and turn them into white to introduce the effect of gaps and incompleteness. We also generate separate binary masks for each shape. Finally, we have an image and a set of masks for each shape. Figure \ref{fig:sample-data}(b) shows a generated sample for Mask R-CNN.

\subsubsection{Training Set and Validation Set}

We generate $100000$ samples as training data and $10000$ samples as validation data. There are at most $5$ rooms in each sample. Based on our measurements in real world rooms, we vary the length of each edge of each room within $2$ to $8$ meters.


\subsection{PtrNet Architecture and Hyperparameters}

Each of our PtrNet models uses a single LSTM layer with 512 hidden units, attention size 512, Adam optimizer with 0.001 learning rate, batch size of 128, beam width of 4 and maximum gradient norm 5.0. Models are trained for 30K-60K steps depending on convergence.


\subsection{Mask R-CNN Architecture and Hyperparameters}

We train Mask R-CNN on the generated dataset. We use Resnet101 as backbone, batch size 8, SGD momentum optimizer with momentum 0.9, learning rate 0.001, gradient clip norm 5.0. The Resnet101 model is initialized with COCO pre-trained weights. Then, we fine-tune the model on our dataset. The model is trained for 60 epochs, with 100 steps per epoch. For the first 30 epochs, only randomly initialized head layers are trained and all backbone layers are kept frozen. Then, in the next 30 epochs, all layers are fine-tuned.

\subsection{Performance Metric}

We measure the performance of our models based on how similar the shapes of predicted rooms and ground-truth rooms are. For similarity measurement, we use Intersection over Union (IoU) score. It is the ratio of the common area and the combined area of predicted room and ground-truth room. IoU values can range from 0 to 1. Higher IoU means higher similarity between prediction and ground-truth. We compare the performance of our different models by mean IoU score. Mean IoU score of a model is the sum of IoU scores of all predicted rooms divided by the total number of predicted rooms.

\subsection{Testing on Real Environment}

We test our model on $10$ indoor scenes. The 3D meshes are collected from those scenes using Google Tango. We use Asus Zenfone AR powered by Google's Project Tango to collect 3D meshes. Then, we sample $3 \times 10^6$ points from each 3D mesh to get 3D point-clouds. The walls are extracted using histogram approach. From the extracted walls, rooms are identified using our trained PtrNet models and Mask R-CNN model.

\section{Evaluation}\label{sec:eval}

In this section we measure the capability of our approaches. We first evaluate them on generated dataset. Then we evaluate them on real world point-clouds data. We also show some example results of our models.

\subsection{Overall Performance on Generated Dataset}

First we measure performance of our models in the validation set of our generated dataset. In Table \ref{tab:iou-vs-len} we show how the performance of PtrNet changes with different ordering schemes and input length. We vary the length of input sequence $P_n$ to 100, 200, 300 and 400. Here, the mean IoU of PtrNet-TrueSort varies from 0.946 to 0.970. The mean IoU of PtrNet-Random varies from 0.553 to 0.840. And, the mean IoU of PtrNet-PseudoSort varies from 0.711 to 0.892. PtrNet-TrueSort outperforms every other sorting schemes. It shows that ordering indeed has a significant effect on PtrNet's performance. In addition, the mean IoU of PtrNet-PseudoSort increases along with input length upto 300, then it decreases at input length 400.

\begin{table}[!ht]
    \centering
    \begin{tabular}{|c|c|c|c|c|} 
    \hline
    \multirow{2}{*}{Sorting Schemes} & \multicolumn{4}{c|}{Input Length} \\\cline{2-5}
     & 100 & 200 & 300 & 400 \\\hline
    PtrNet-TrueSort & 0.970 & 0.953 & 0.945 & 0.946 \\\hline
    PtrNet-Random & 0.553 & 0.558 & 0.602 & 0.840 \\\hline
    PtrNet-PseudoSort & 0.711 & 0.803 & 0.892 & 0.843 \\\hline
    \end{tabular}
    \caption{PtrNet: Mean IoU vs. input length}
    \label{tab:iou-vs-len}
\end{table}

We also see that a simple ordering like PseudoSort has 44\%-48\% performance gain from PtrNet-Random. It shows, the closer the ordering to TrueSort, the better the performance. We select best model, PtrNet-PseudoSort with 300 input points for real-world evaluation. We also measure how Mask R-CNN performs in generated dataset. This is reported in Table \ref{tab:iou-best}. Both approaches perform almost similarly on generated dataset.

\begin{table}[!ht]
    \centering
    \begin{tabular}{|c|c|} 
    \hline
    Model & Mean IoU \\\hline
    PtrNet-PseudoSort (300) & 0.8921 \\\hline
    Mask R-CNN & 0.8919 \\\hline
    \end{tabular}
    \caption{Performance in generated data}
    \label{tab:iou-best}
\end{table}

\subsection{Shape Based Performance on Generated Dataset}

We measure how our proposed models are able to detect different shapes. For this, we keep track of IoU scores of different shapes of rooms. This is reported in Table \ref{tab:iou-vs-shape}. Mask R-CNN and PtrNet-PseudoSort perform almost similarly in extracting all kinds of shapes in generated dataset. 

\begin{table}[!ht]
    \centering
    \begin{tabular}{|c|c|c|} 
    \hline
    \multirow{2}{*}{Models} & \multicolumn{2}{c|}{Shapes} \\\cline{2-3}
     & Rectangular & Rectilinear\\\hline
    PtrNet-TrueSort & 0.927 & 0.949\\\hline
    PtrNet-Random & 0.650 & 0.467\\\hline
    PtrNet-PseudoSort & 0.862 & 0.856\\\hline
    Mask R-CNN & 0.910 & 0.823\\\hline
    \end{tabular}

    \bigskip

    \begin{tabular}{|c|c|c|} 
    \hline
    \multirow{2}{*}{Models} & \multicolumn{2}{c|}{Shapes} \\\cline{2-3}
     & Curved & Non-right Angle \\\hline
    PtrNet-TrueSort & 0.977 & 0.937 \\\hline
    PtrNet-Random & 0.640 & 0.587 \\\hline
    PtrNet-PseudoSort & 0.919 & 0.884 \\\hline
    Mask R-CNN & 0.917 & 0.903 \\\hline
    \end{tabular}
    \caption{Mean IoU vs. room shape}
    \label{tab:iou-vs-shape}
\end{table}

\subsection{Performance on Real World Dataset}

Finally, we take two of our best models, PtrNet-PseudoSort (300) and Mask R-CNN to evaluate them on real world data. For this, we collect Point-cloud from 10 indoor scenes. The indoor scenes include classrooms, households, office spaces. To show our models can recognize rooms of different shapes, we change the shapes of some rooms by hanging sheets vertically from ceiling to floor. Table \ref{tab:iou-real} shows, mean IoU scores of our two approaches in real dataset.

\begin{table}[!ht]
    \centering
    \begin{tabular}{|c|c|} 
    \hline
    Model & Mean IoU \\\hline
    PtrNet-PseudoSort (300) & 0.560 \\\hline
    Mask R-CNN & 0.850 \\\hline
    \end{tabular}
    \caption{Performance in real world data}
    \label{tab:iou-real}
\end{table}

\begin{figure*}[!ht]
    \captionsetup{position=top}
    \captionsetup[subfigure]{labelformat=empty}
    \centering
        \subfloat[Point-cloud]{\includegraphics[width=0.2\textwidth]{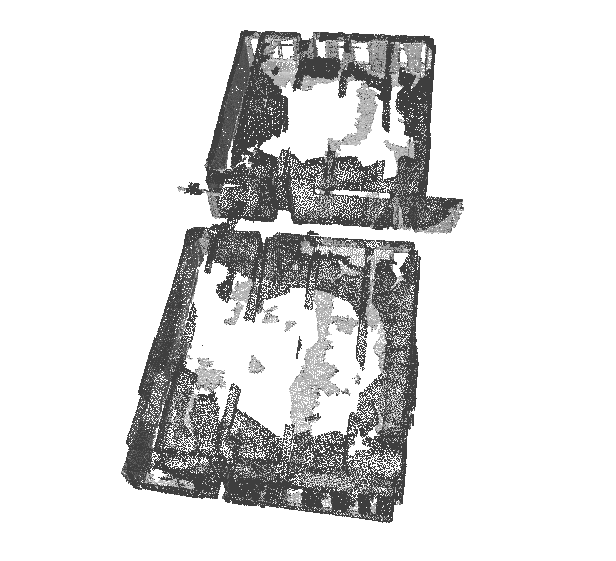}}
        \subfloat[Extracted walls]{\includegraphics[width=0.2\textwidth]{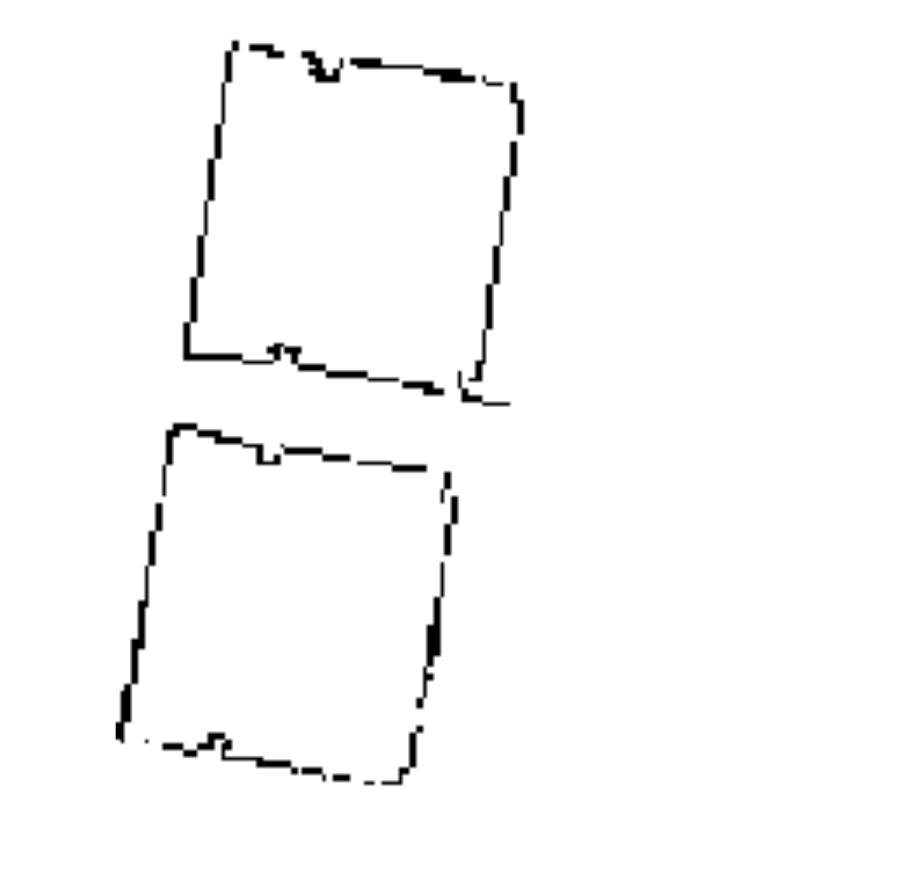}}
        \subfloat[PtrNet]{\includegraphics[width=0.2\textwidth]{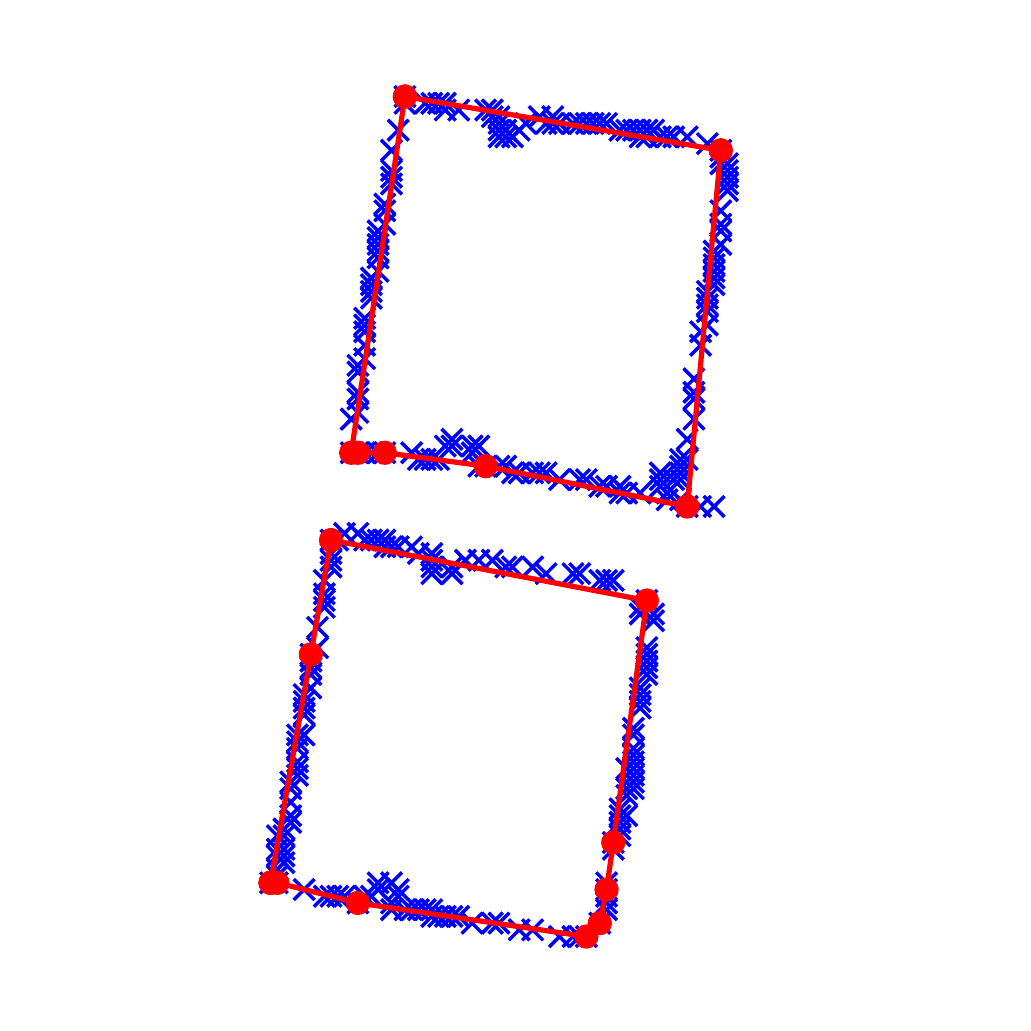}}
        \subfloat[Mask R-CNN]{\includegraphics[width=0.2\textwidth]{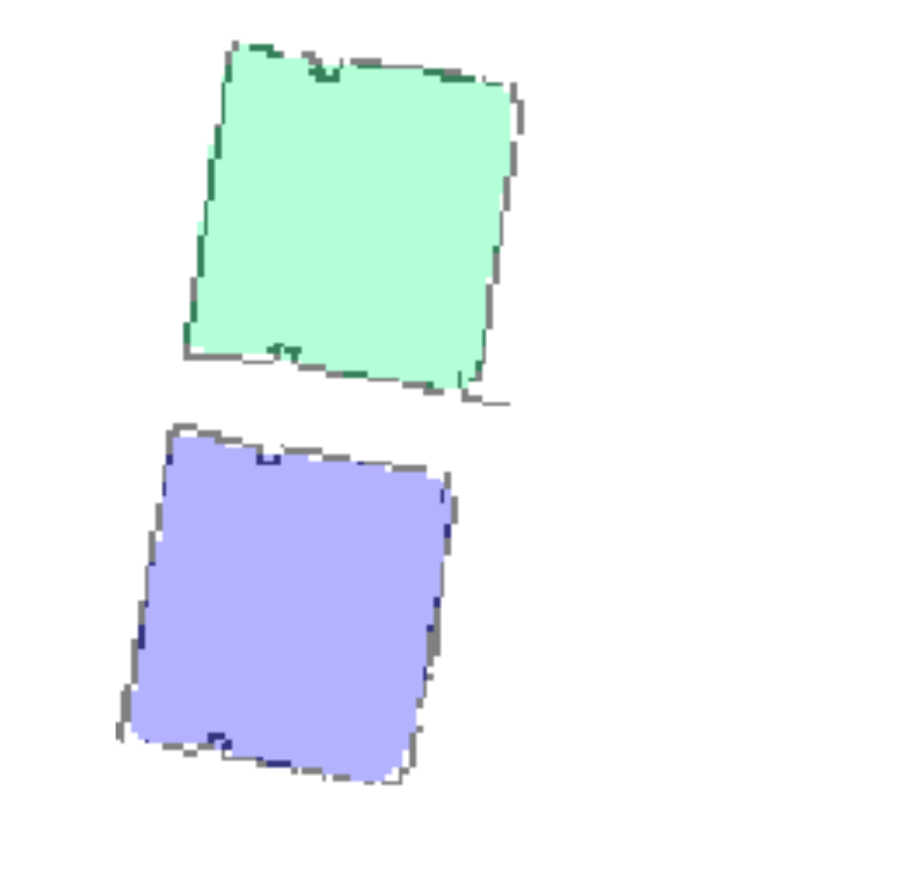}}
        \subfloat[Ground Truth]{\includegraphics[width=0.2\textwidth]{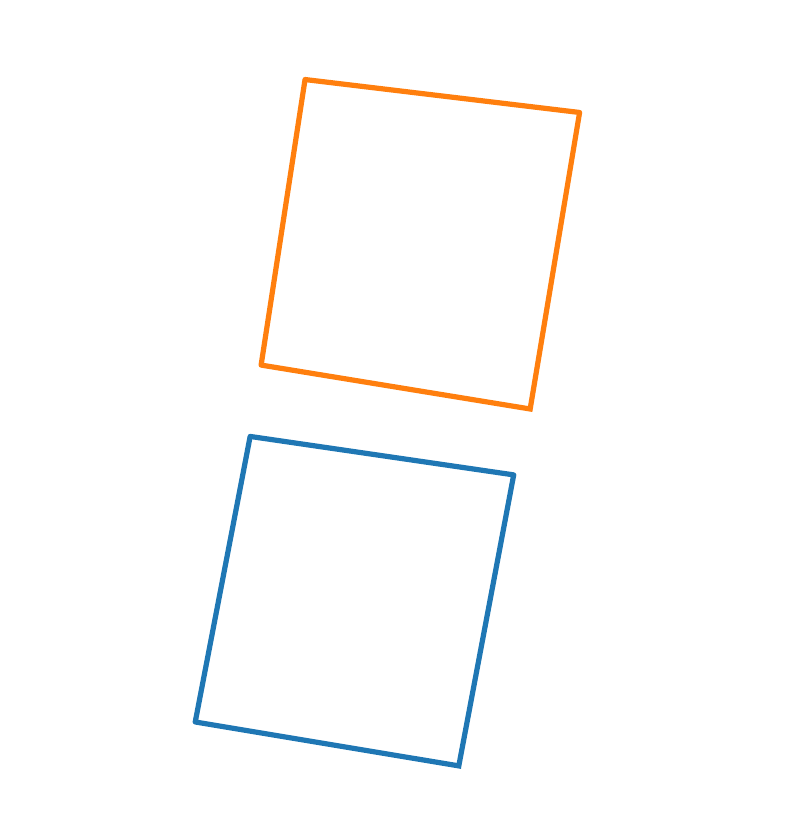}}
        \\
        \subfloat[Point-cloud]{\includegraphics[width=0.2\textwidth]{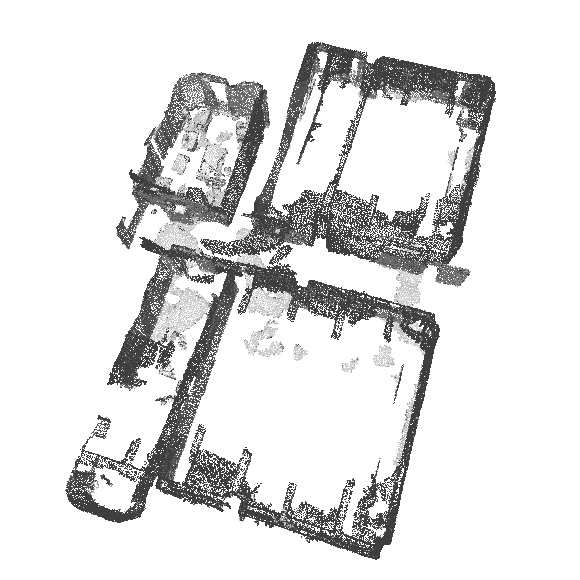}}
        \subfloat[Extracted walls]{\includegraphics[width=0.2\textwidth]{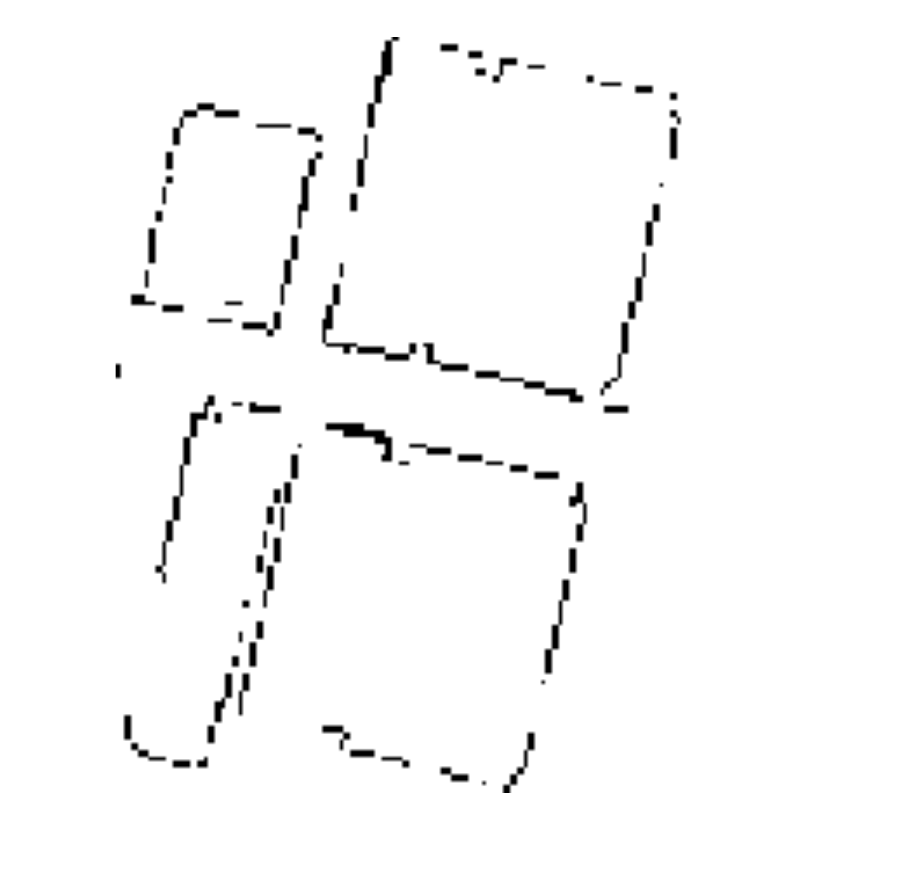}}
        \subfloat[PtrNet]{\includegraphics[width=0.2\textwidth]{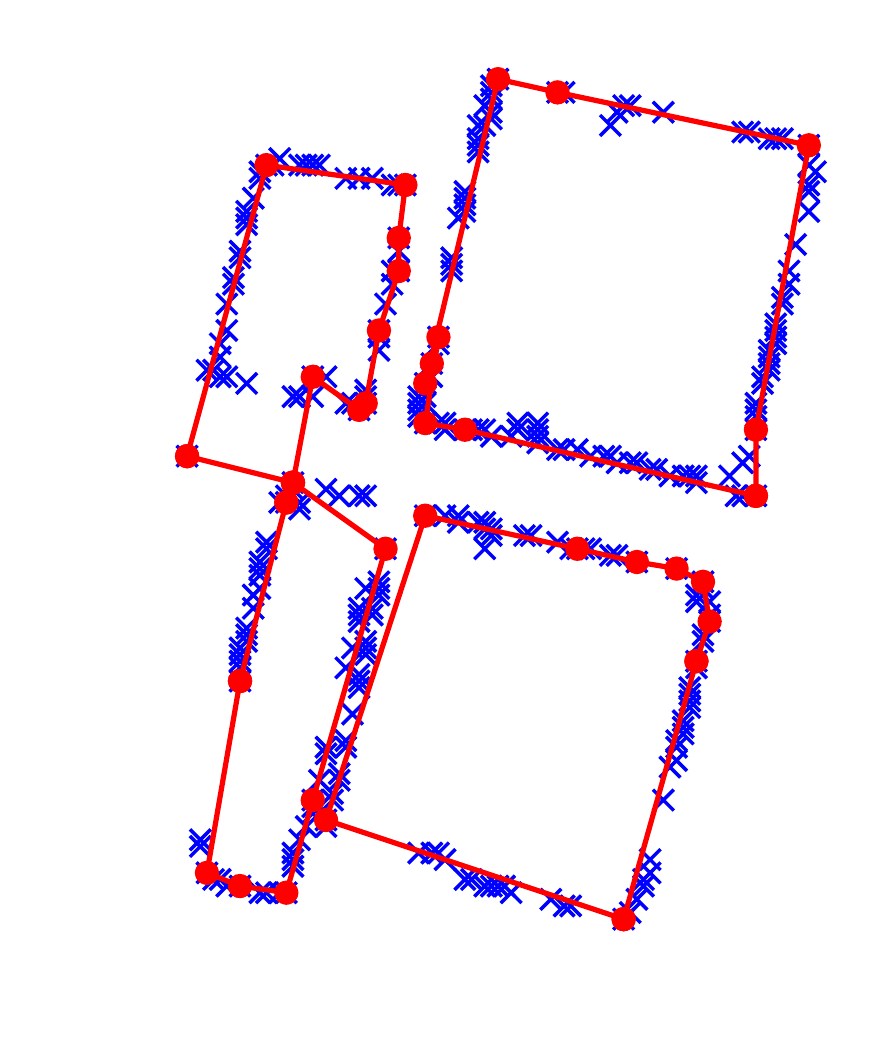}}
        \subfloat[Mask R-CNN]{\includegraphics[width=0.2\textwidth]{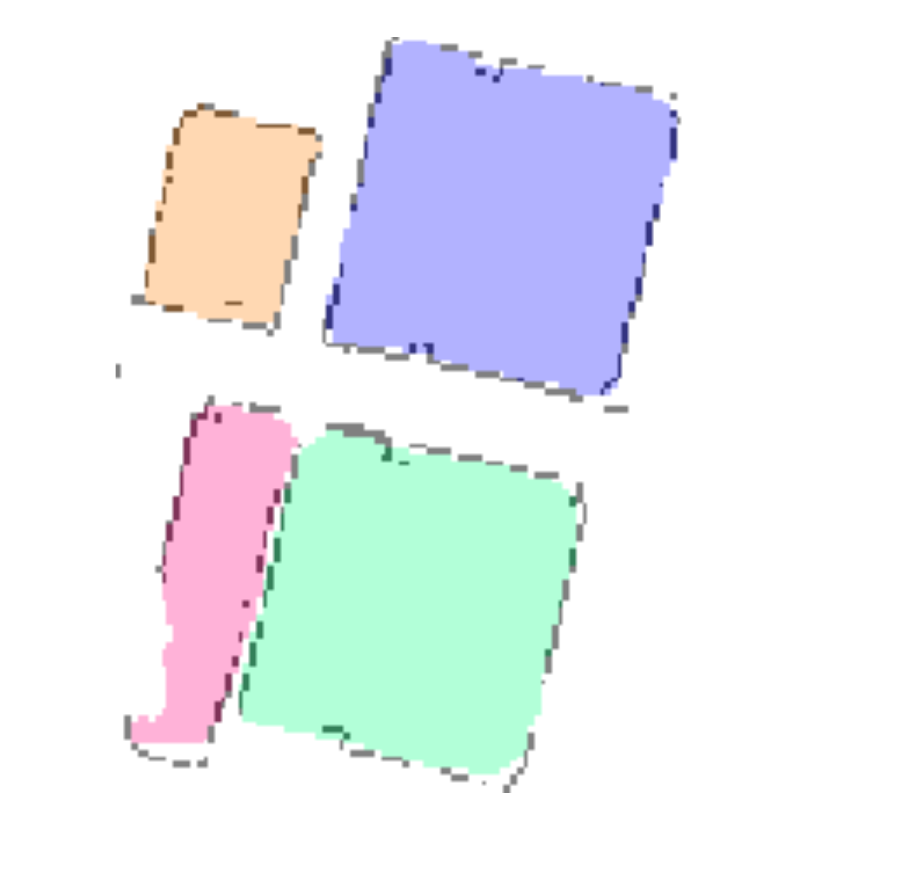}}
        \subfloat[Ground Truth]{\includegraphics[width=0.2\textwidth]{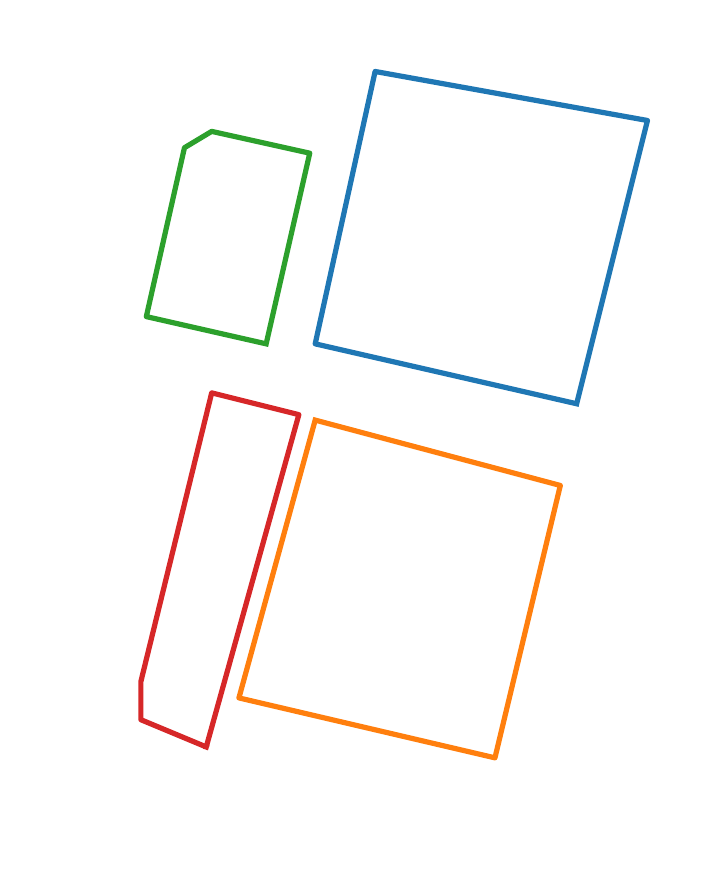}}
        \\
        \subfloat[Point-cloud]{\includegraphics[width=0.2\textwidth]{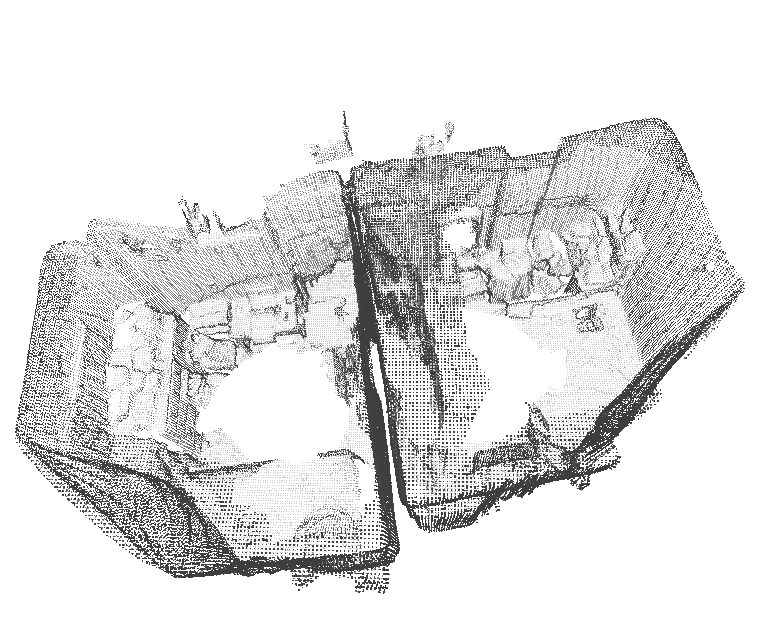}}
        \subfloat[Extracted walls]{\includegraphics[width=0.2\textwidth]{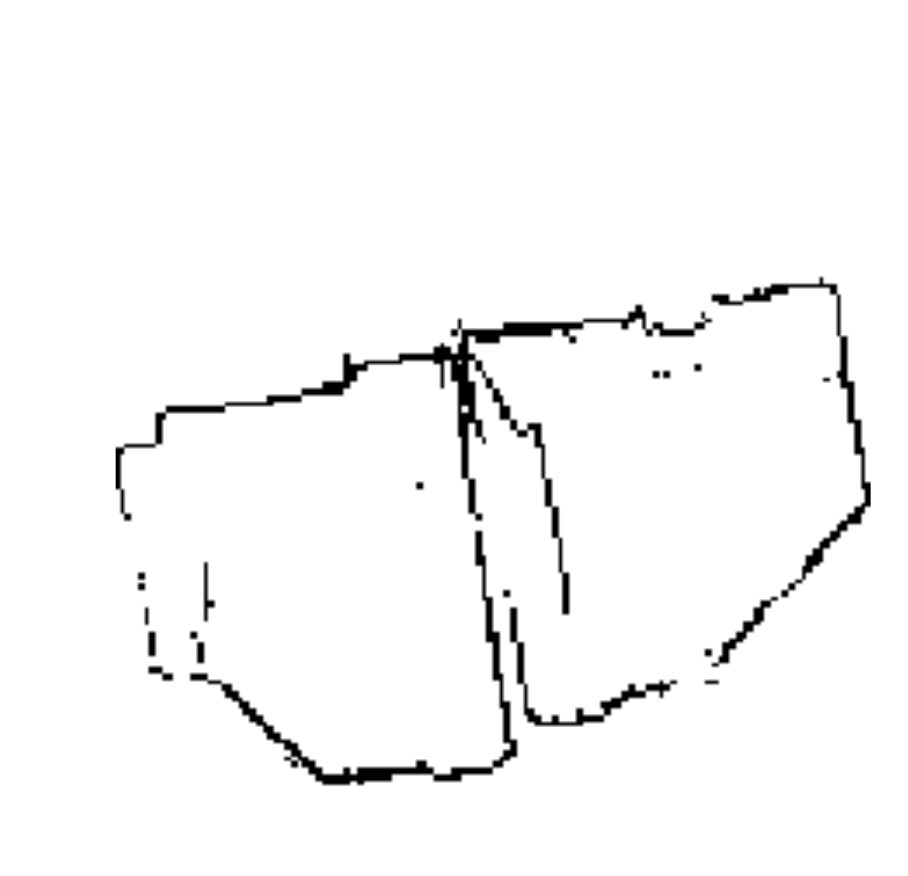}}
        \subfloat[PtrNet]{\includegraphics[width=0.2\textwidth]{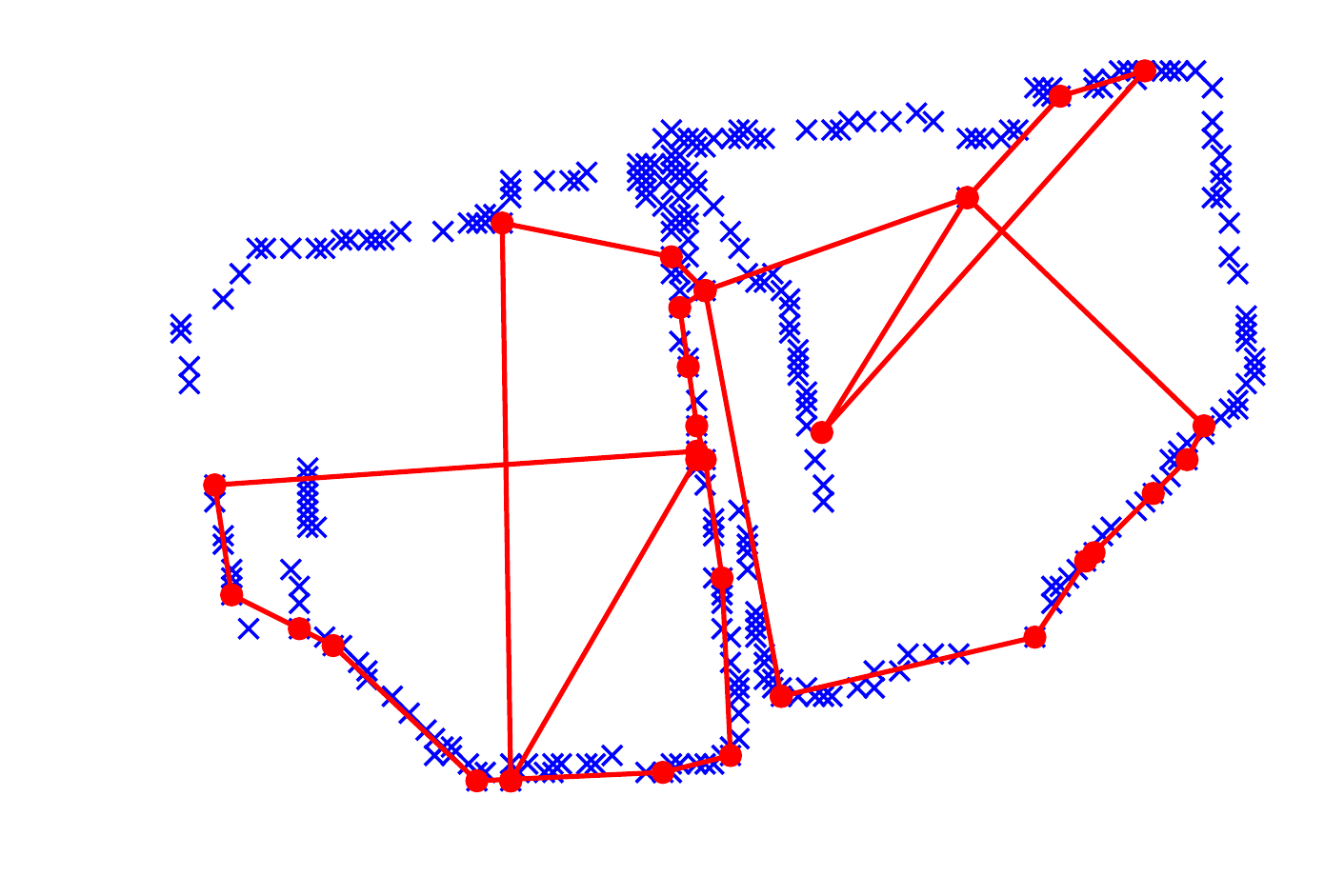}}
        \subfloat[Mask R-CNN]{\includegraphics[width=0.2\textwidth]{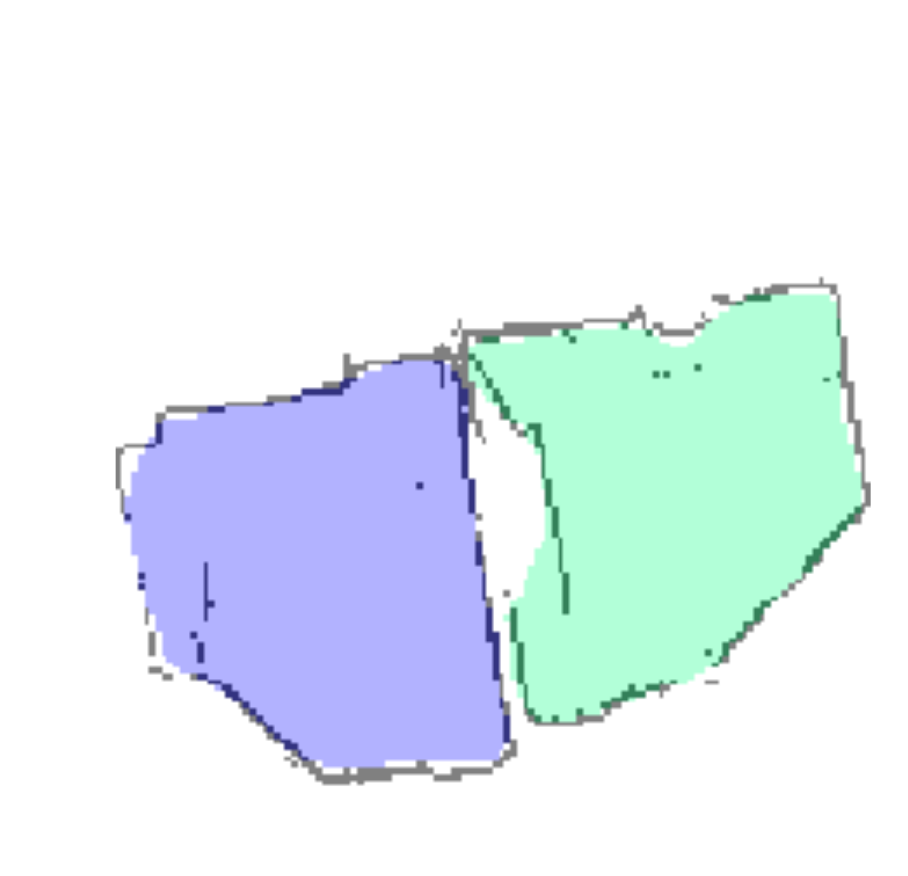}}
        \subfloat[Ground Truth]{\includegraphics[width=0.2\textwidth]{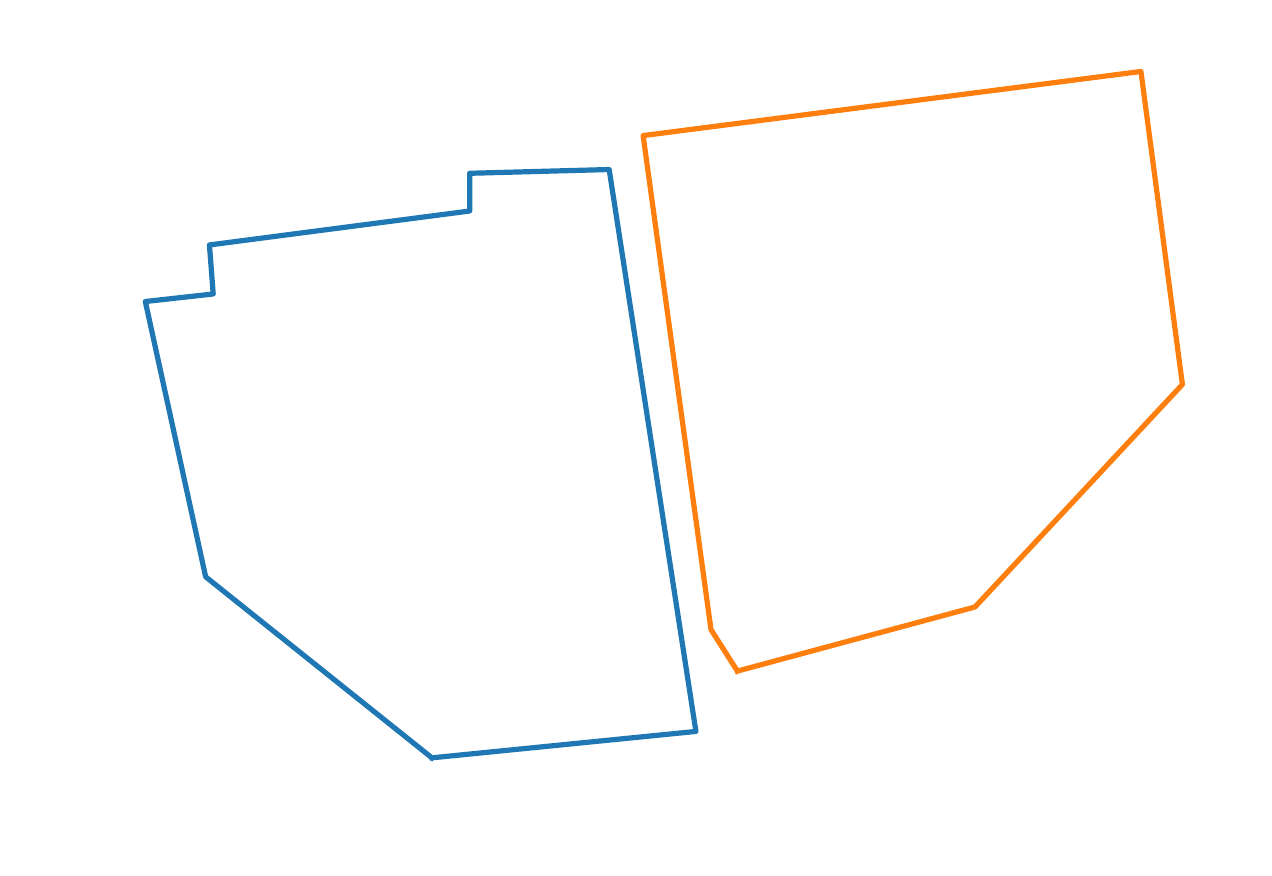}}
        \\
        \subfloat[Point-cloud]{\includegraphics[width=0.2\textwidth]{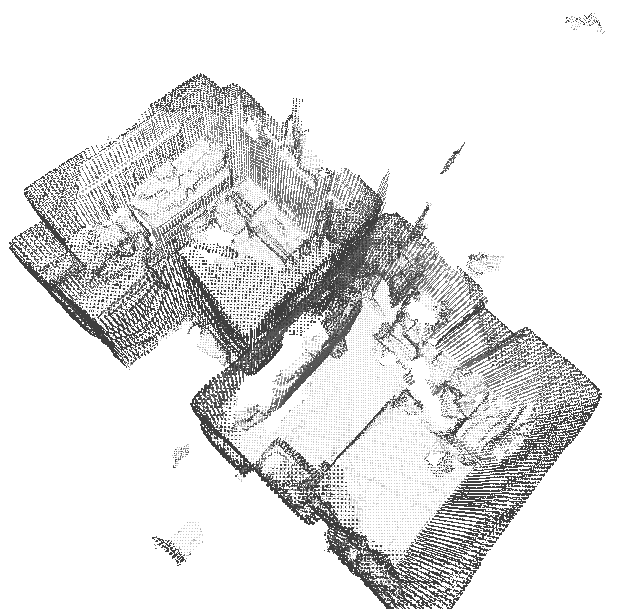}}
        \subfloat[Extracted walls]{\includegraphics[width=0.2\textwidth]{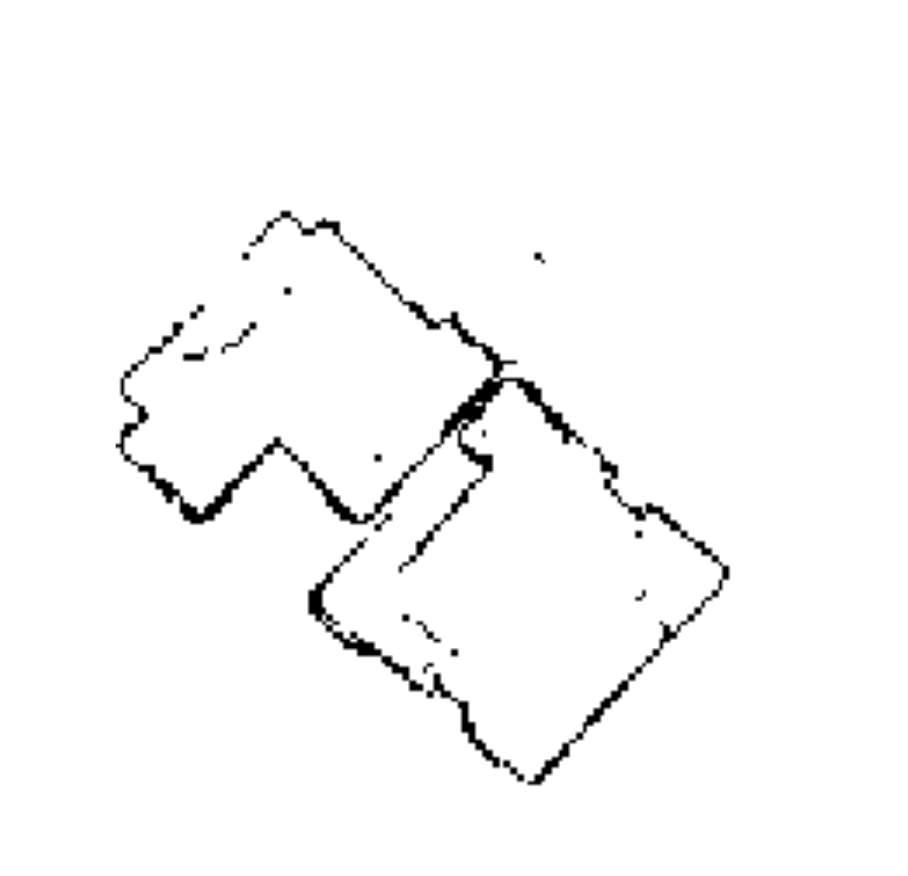}}
        \subfloat[PtrNet]{\includegraphics[width=0.2\textwidth]{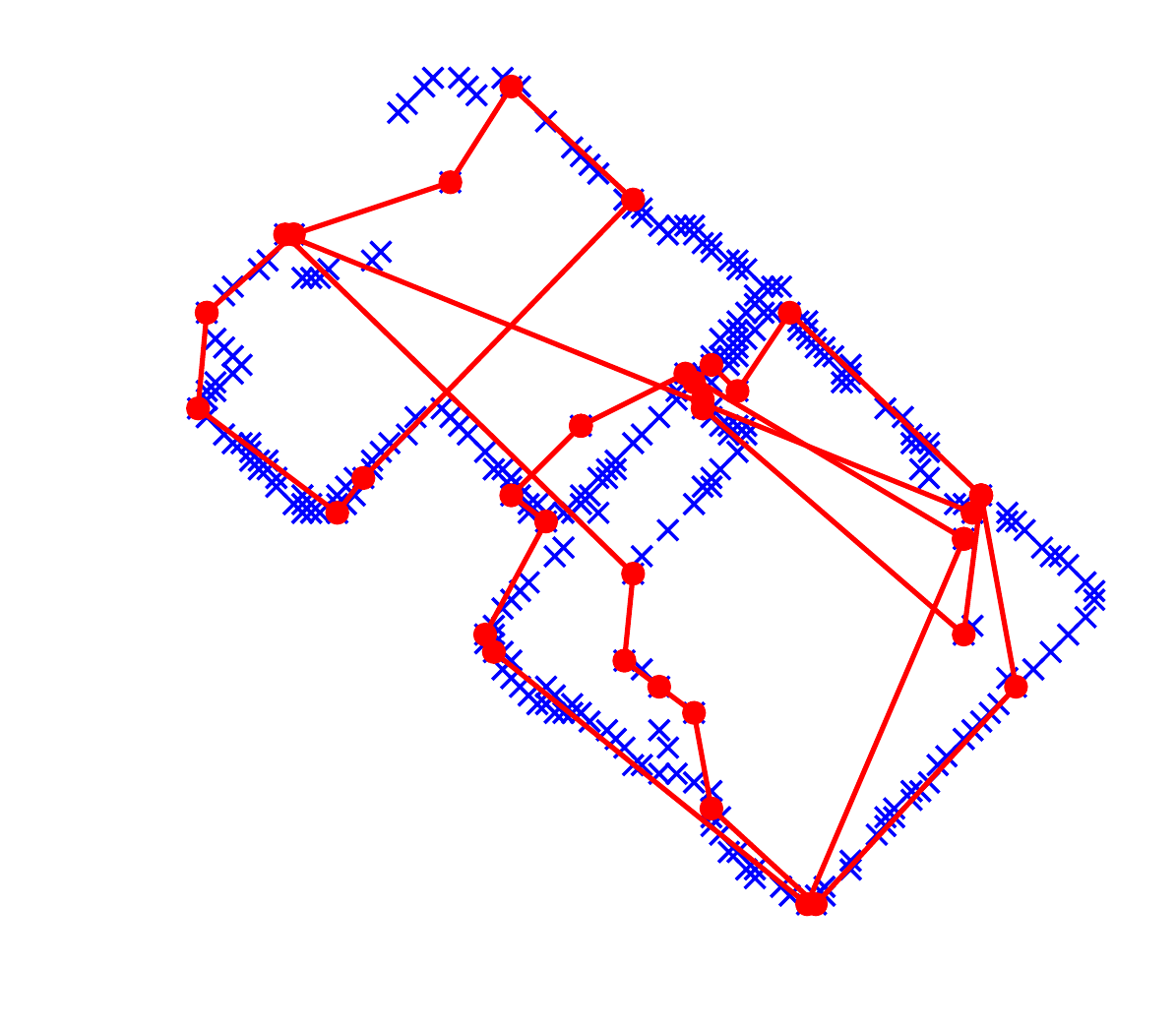}}
        \subfloat[Mask R-CNN]{\includegraphics[width=0.2\textwidth]{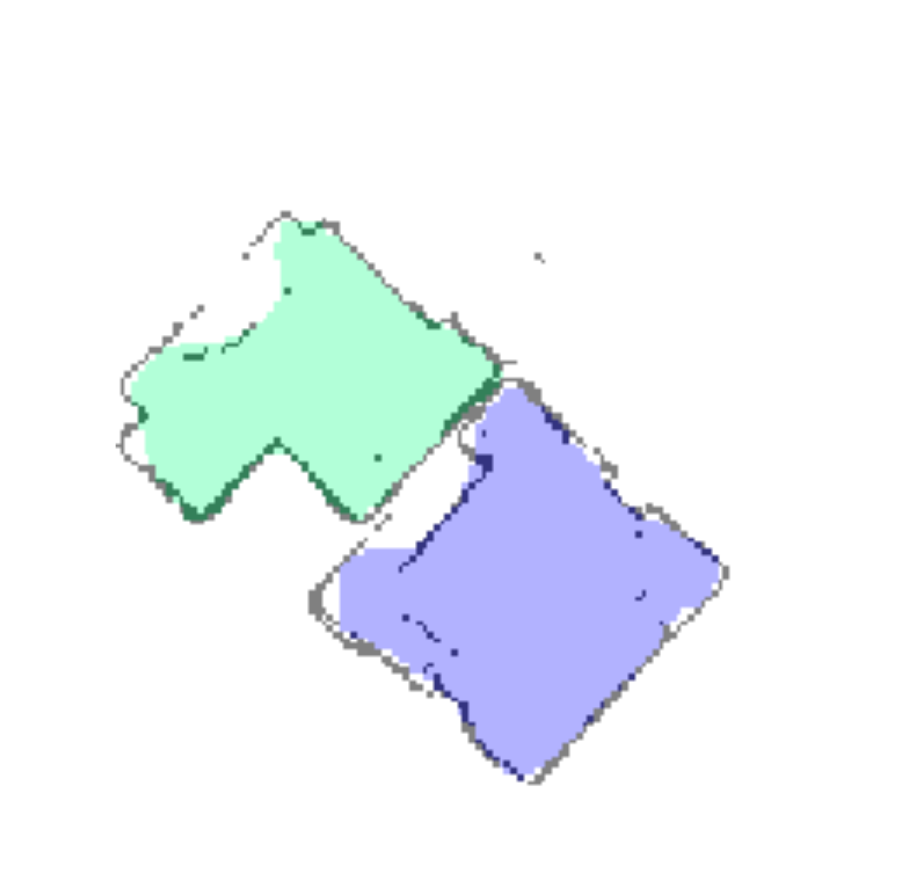}}
        \subfloat[Ground Truth]{\includegraphics[width=0.2\textwidth]{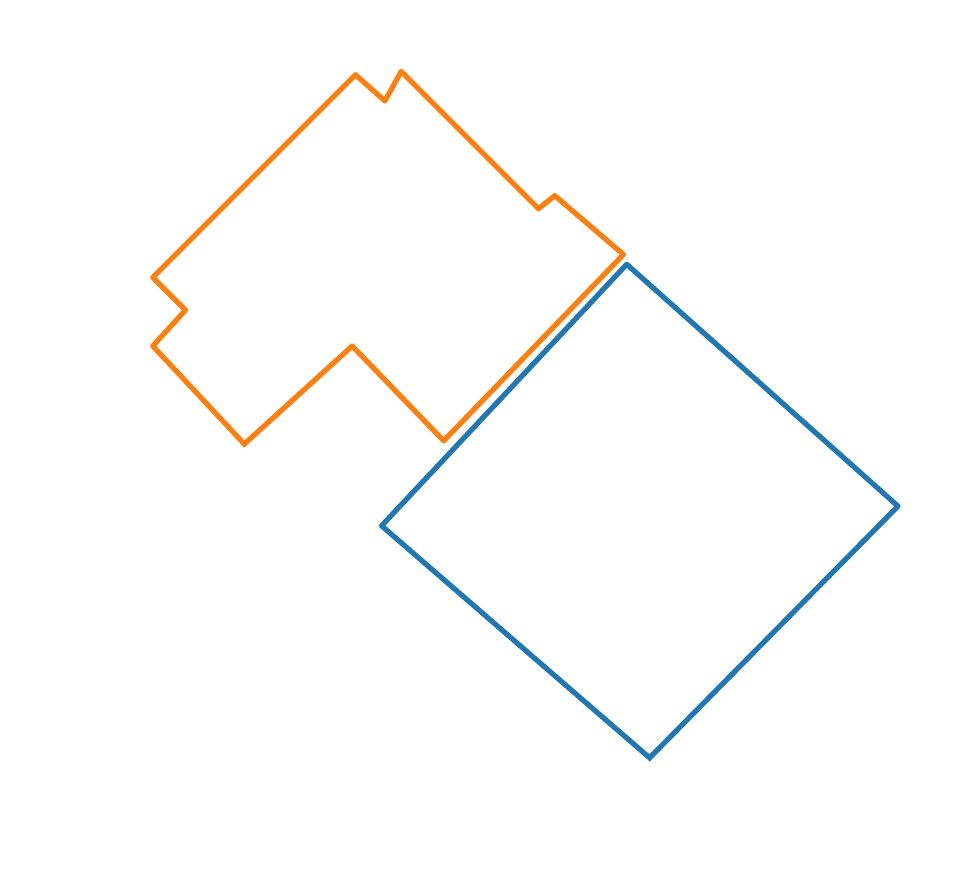}}
        
        \caption{Real world indoor layout results}
    \label{fig:real}
\end{figure*}

Table \ref{tab:iou-real} shows, Mask R-CNN has an 52\% performance gain from PtrNet in real world dataset. We show some results of our proposed systems in Figure \ref{fig:real}. In the first example, walls are extracted almost perfectly and clutters are completely excluded. Both PtrNet and Mask R-CNN detect rooms accurately. In the second example, again clutters are mostly excluded by the wall detection module. But there are visible gaps in extracted walls due to windows and doors. Especially, there is a large gap in the lower left room due to the existence of a large window. Both Mask R-CNN and PtrNet detect three rooms accurately. But PtrNet over estimates the area of the top-left room. And Mask R-CNN suffers slightly due to the large gap in the bottom-left room. In the third example, there is a large wardrobe in the right room. The wall detection module fails to exclude it completely. Also, there are some other clutters still present after wall extraction. Although, Mask R-CNN is able to detect rooms, but PtrNet performs poorly on this example. Again, in the fourth example, there are still some background clutters. So, again PtrNet fails completely even though Mask R-CNN detects both the rooms. So, it is evident from these examples that Mask R-CNN can detect rooms reasonably well compared to PtrNet.

\section{Conclusion}\label{sec:conclu}

In this paper, we introduce an indoor mapping system using smart hand-held devices. First, we capture the indoor scene in the form of a 3D mesh or a point-cloud generated by a simple scan of modern mobile device. Then, we use 2D histogram based approach to detect positions of walls from the points-clouds. Finally, we use two paradigms of deep learning frameworks: one from geometric learning and the other from image segmentation, to detect rooms from walls. Since deep learning models require huge amounts of data to train and currently no publicly available dataset is available for this purpose, we also develop a method to generate millions of samples for training these networks. Though our models are trained on synthetic data, our experiments show satisfactory results in real-world environments. This work is a promising start towards learning based spatial object detection. In future work, we will try to show applicability of this approach in detecting other spatial objects e.g. doors, windows, staircase. 


\bibliographystyle{ACM-Reference-Format}
\bibliography{room}

\end{document}